\documentclass[10pt,twocolumn,letterpaper]{article}

\usepackage{iccv}
\usepackage{times}
\usepackage{epsfig}
\usepackage{graphicx}
\usepackage{amsmath}
\usepackage{amssymb}
\usepackage{booktabs}
\makeatletter
\@namedef{ver@everyshi.sty}{}
\makeatother
\usepackage{tikz}
\usepackage{comment}
\usepackage{array}
\usepackage{multirow}
\usepackage{tabularx}
\usepackage{caption}
\usepackage{subcaption}
\usepackage{arydshln}

\usepackage[breaklinks=true,bookmarks=false]{hyperref}

\iccvfinalcopy 

\def\iccvPaperID{****} 

\ificcvfinal\fi

\usepackage{xspace}
\usepackage[export]{adjustbox}
\usepackage[abs]{overpic}
\usetikzlibrary{arrows.meta}

\usepackage{bold-extra}
\usepackage{cite}

\usepackage{tikz}
\usetikzlibrary{shapes.geometric,positioning,backgrounds,arrows,patterns,calc,fit}
\usepackage{pgfplots}
\pgfplotsset{compat=1.15}

\makeatletter

\definecolor{notetext}{rgb}{0.7,0,0}

\definecolor{ubpubColor}{rgb}{0.43, 0.5, 0.5}
\definecolor{backboneColor}{rgb}{0.423, 0.325, 0.365}
\definecolor{fpnColor}{rgb}{0.255, 0.498, 0.416}

\newcommand{\PAR}[1]{\vskip4pt \noindent {\bf #1~}}
\newcommand{\PARbegin}[1]{\noindent {\bf #1~}}

\newcolumntype{P}[1]{>{\centering\arraybackslash}p{#1}}

\newcommand{\abbrev}{DynaMITe}
\newcolumntype{Y}{>{\centering\arraybackslash}X}

\definecolor{backbone_color}{HTML}{ff6d01}
\definecolor{feature_color}{RGB}{128, 128, 128}
\definecolor{click_sampling_color}{HTML}{9900ff}
\definecolor{transformer_color}{HTML}{34a853}

\definecolor{q1c}{HTML}{4285f4}
\definecolor{q2c}{HTML}{4285f4}
\definecolor{qtc}{HTML}{34a853}
\definecolor{qtp1c}{HTML}{ea4335}
\definecolor{bg1c}{HTML}{808080}
\definecolor{bg2c}{HTML}{bebebe}

\definecolor{c_red}{HTML}{ea4335}
\definecolor{c_blue}{HTML}{4285f4}
\definecolor{c_green}{HTML}{34a853}
\definecolor{c_orange}{HTML}{ff6d01}

\begin{document}

\title{\vspace{-15pt}\abbrev: Dynamic Query Bootstrapping for Multi-object Interactive Segmentation Transformer\\[-8pt]}

\author{Amit Kumar Rana$^*$
\and
Sabarinath Mahadevan$^*$
\and
Alexander Hermans
\and
Bastian Leibe\\
\\
RWTH Aachen University, Germany\\
{\tt\small {firstname.lastname}@rwth-aachen.de}\\
\small{\url{https://vision.rwth-aachen.de/dynamite}}
}
\maketitle

\begin{NoHyper}
\def\thefootnote{*}\footnotetext{Equal contribution.}
\addtocounter{footnote}{-1}%
\end{NoHyper}

\ificcvfinal\thispagestyle{empty}\fi

\begin{abstract}
Most state-of-the-art instance segmentation methods rely on large amounts of pixel-precise ground-truth annotations for training, which are expensive to create. Interactive segmentation networks help generate such annotations based on an image and the corresponding user interactions such as clicks. Existing methods for this task can only process a single instance at a time and each user interaction requires a full forward pass through the entire deep network. We introduce a more efficient approach, called DynaMITe, in which we represent user interactions as spatio-temporal queries to a Transformer decoder with a potential to segment multiple object instances in a single iteration. Our architecture also alleviates any need to re-compute image features during refinement, and requires fewer interactions for segmenting multiple instances in a single image when compared to other methods.
DynaMITe achieves state-of-the-art results on multiple existing interactive segmentation benchmarks, and also on the new multi-instance benchmark that we propose in this paper.
\end{abstract}


\vspace{-4pt}
\section{Introduction}
\label{sec:intro}
\vspace{-4pt}

Interactive segmentation algorithms enable a user to annotate the objects of interest within a given image with the help of user interactions such as scribbles and clicks. Such algorithms have several advantages compared to fully-automatic segmentation methods, since they enable a user to select and iteratively refine the objects of interest. Existing interactive segmentation methods~\cite{mahadevan18bmvc, SofiiukArxiv21, Sofiiukf20CVPR, Chen22CVPR,liu2022eccv} formulate this task as a binary instance segmentation problem, where the single object of interest can be segmented and corrected using user clicks. 

Most of these approaches use deep neural networks to generate the image features that are conditioned on the user clicks and previous predictions, and they require the image level features to be re-computed for every user interaction. While such a design has been proven to be effective, the runtime
for processing each interaction is proportional to the size of the feature extractor used,
since a forward pass through the network is needed per interaction ~\cite{mahadevan18bmvc, SofiiukArxiv21, Sofiiukf20CVPR, Chen22CVPR,liu2022eccv}. 
Hence, these methods often have to limit their network sizes in order to achieve a good runtime performance and are thus not scalable in this respect. 

\begin{figure*}
\centering

\def\imwidth{33.2mm}

\begin{tikzpicture}[font=\scriptsize]

\node[inner sep=0pt] (t1) {\includegraphics[width=\imwidth]{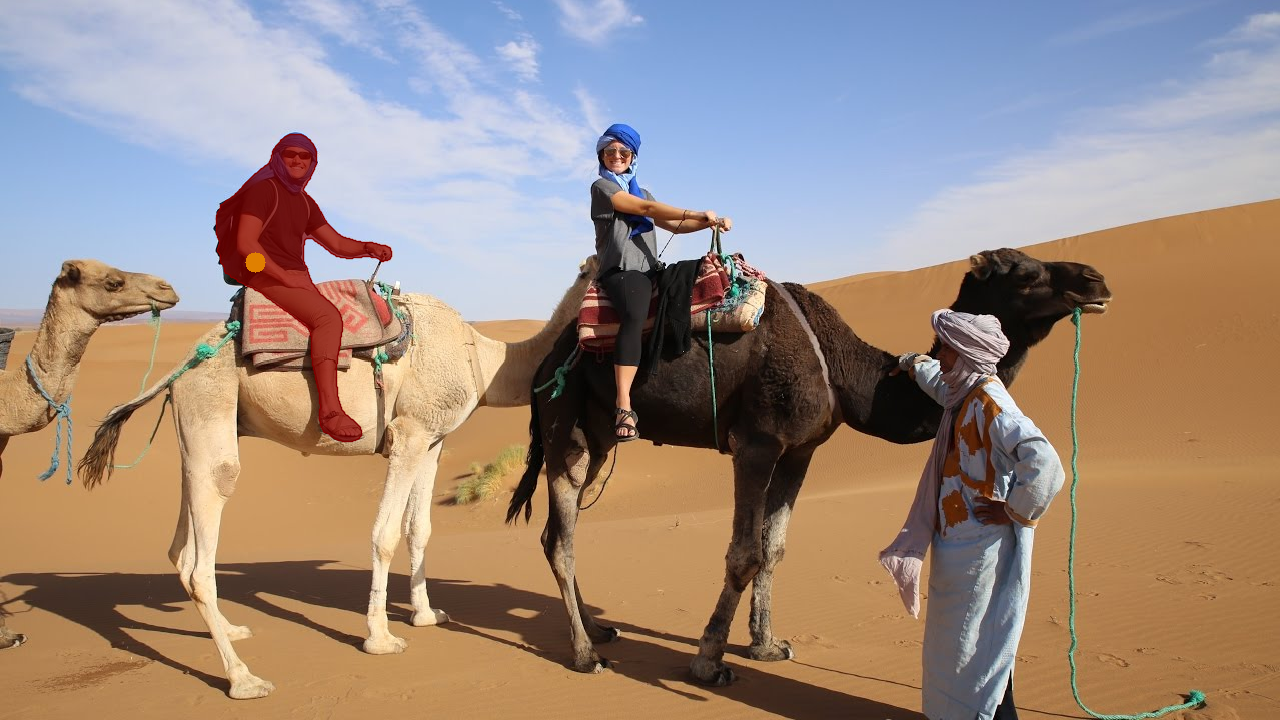}};

\node[inner sep=0pt, right = 1mm of t1.east, anchor=west
] (t2) {\includegraphics[width=\imwidth]{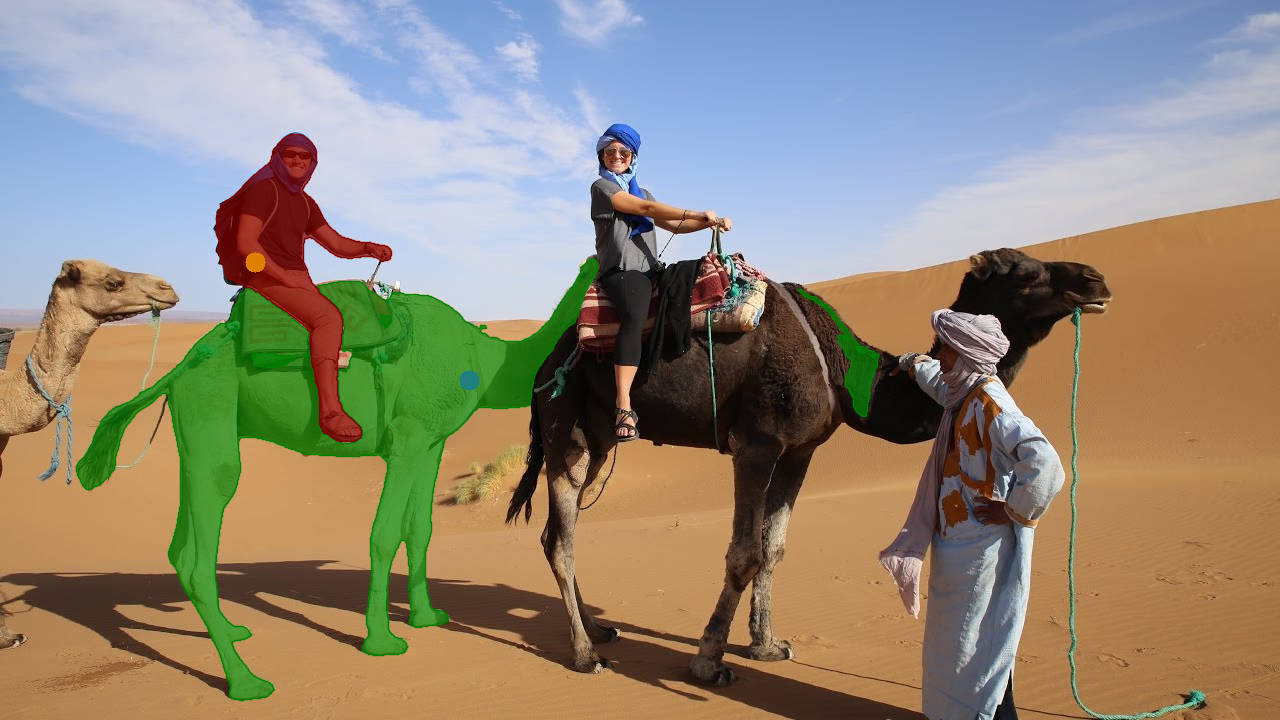}};

\node (fptext) [above = 0.75mm of t2.north, anchor=south, align=center, draw=c_red, fill=c_red!50, thick, rounded corners] {False Positive Area};

\node (box) [anchor = center, rectangle, draw, inner sep=1pt, very thick, draw=c_red, fit= {($(t2.center)+(3mm,3mm)$)($(t2.center)+(8mm, -3mm)$)}]{};
\draw [-latex, densely dashed, c_red, thick] (fptext.-45) to [out=-90,in=-250] (box.north);

\node[inner sep=0pt, right = 1mm of t2.east, anchor=west
] (t3) {\includegraphics[width=\imwidth]{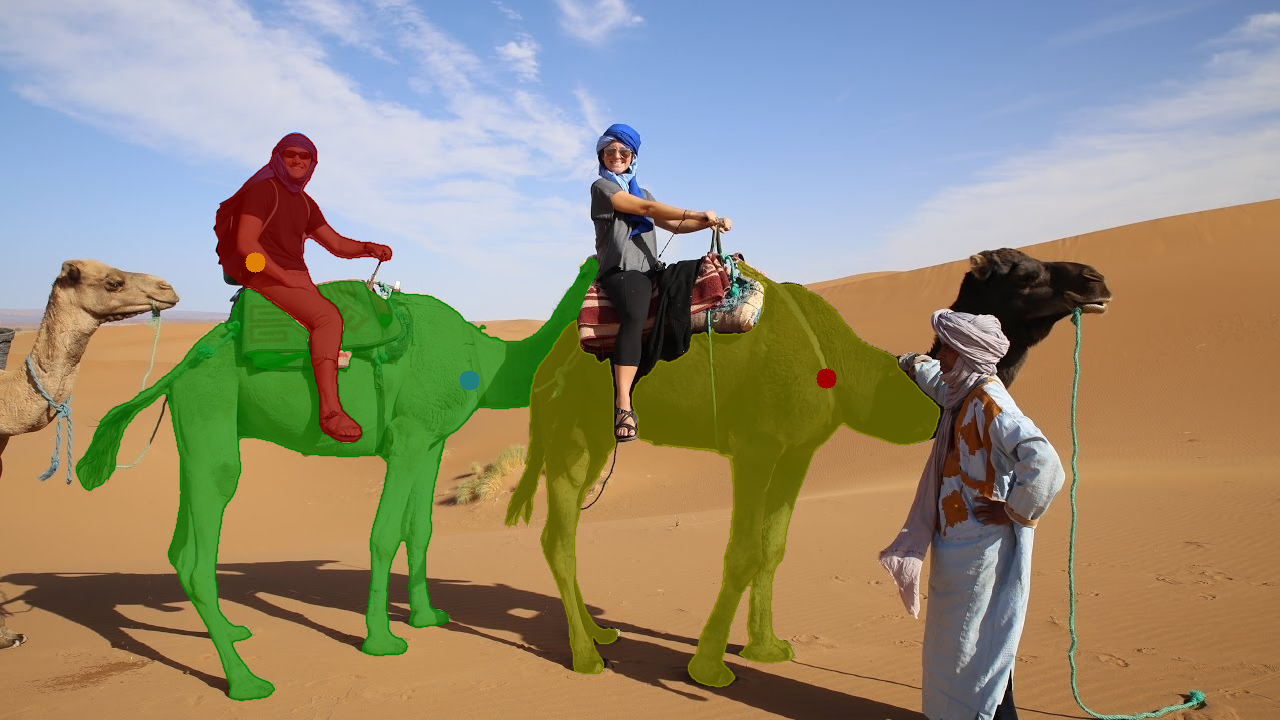}};

\node (correctiontext) [above = 0.75mm of t3.north, anchor=south, align=center, draw=c_green, fill=c_green!50, thick, rounded corners] {Corrective Click};

\node (click1) [anchor = center, circle,draw, inner sep=1.2pt, fill=c_green!60, draw=c_green, thick] at ($(t3.center)+(4.9mm, -0.55mm)$) {};
\draw [-latex, densely dashed, c_green, thick] (correctiontext.-45) to [out=270,in=100] (click1);

\node[inner sep=0pt, right = 1mm of t3.east, anchor=west
] (t4) {\includegraphics[width=\imwidth]{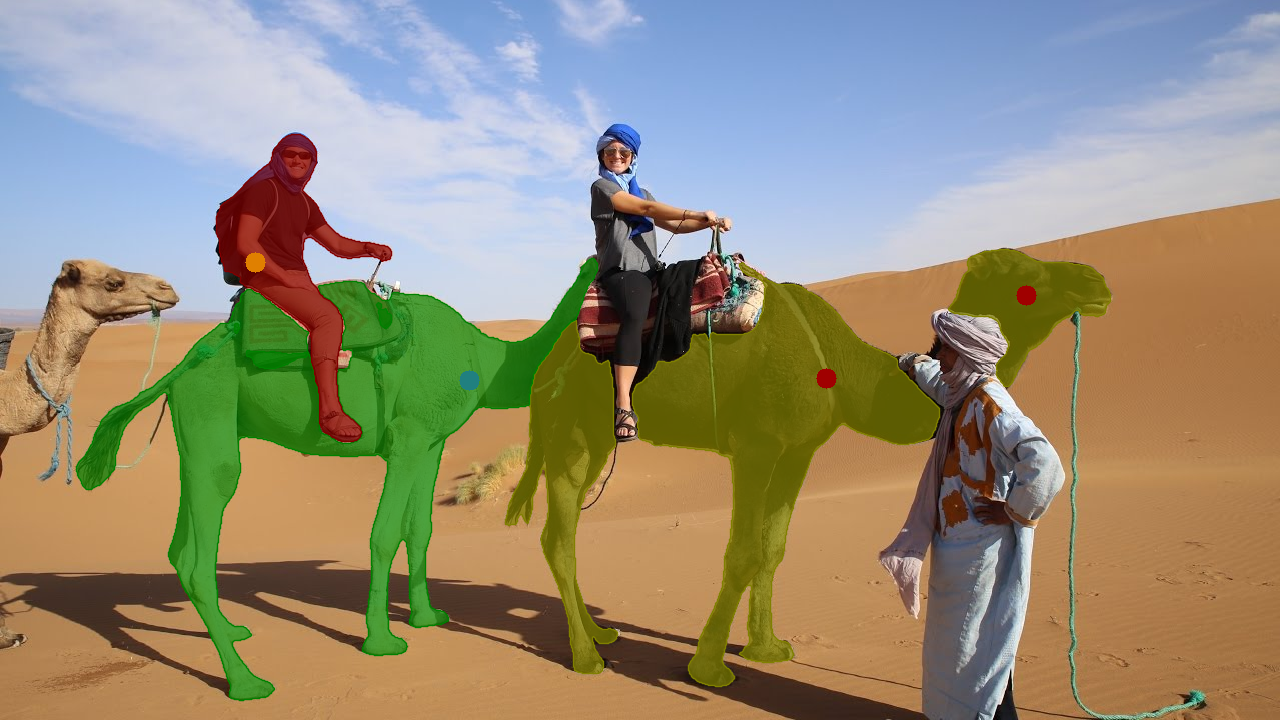}};

\node (additiontext) [above = 0.75mm of t4.north, anchor=south, align=center, draw=c_orange, fill=c_orange!50, thick, rounded corners] {Additional Click};

\node (click2) [anchor = center, circle,draw, inner sep=1.2pt, fill=c_orange!60, draw=c_orange, thick] at ($(t4.center)+(9.9mm, 1.5mm)$) {};
\draw [-latex, densely dashed, c_orange, thick] (additiontext.-25) to [out=270,in=110] (click2);

\node[inner sep=0pt, right = 5mm of t4.east, anchor=west
] (t5) {\includegraphics[width=\imwidth]{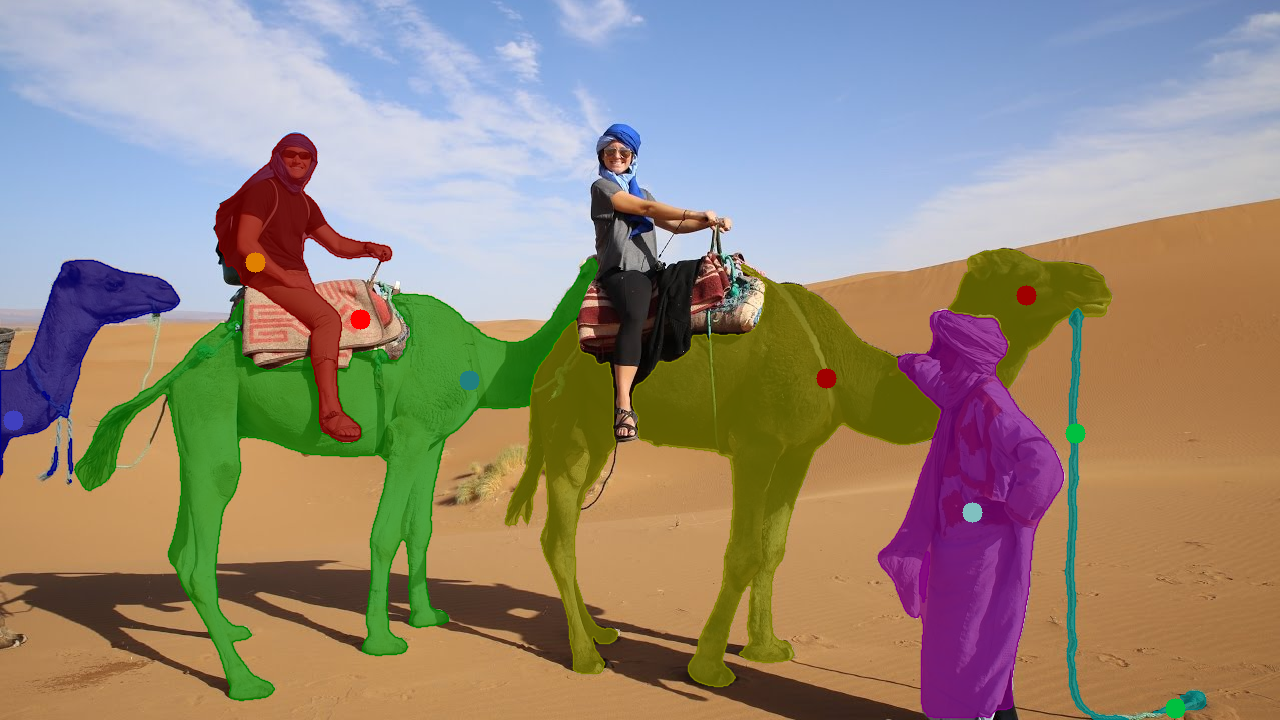}};

\draw[dotted, shorten <=4pt, shorten >=3pt, thick] (t4.east) -- (t5.west);

\node[above = 0.75mm of t5.north, anchor=south, align=center] {Final Segmentation};

\draw[-latex, ultra thick, draw=lightgray] ($(t1.south)-(0,2.5mm)$) -- ($(t5.south)-(0,2.5mm)$) node[pos=0.5, fill=white] {Click-based Refinement Process} ;

\end{tikzpicture}
  \caption{\textbf{\abbrev} processes multiple instances at once and models the background jointly. In this example, the false positive region on the camel in the second image is corrected automatically when the user chooses to segment it as foreground. \abbrev\ is also able to correctly segment tiny structures, such as the camel's leash in the final segmentation mask.
  }
  \label{fig:teaser}
\end{figure*}
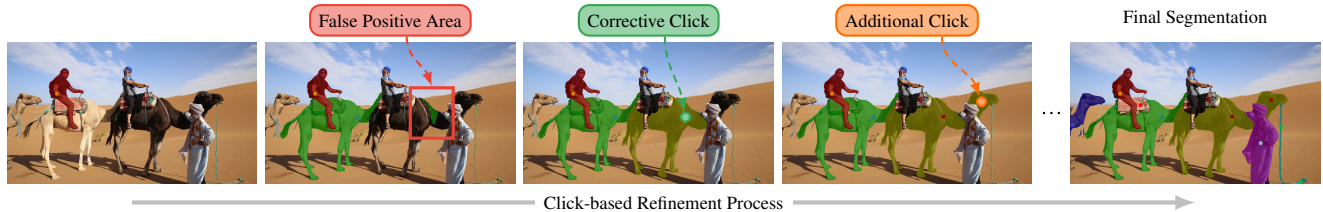

In addition, the design decision to model interactive segmentation as a binary segmentation problem forces existing methods to approach multi-instance segmentation tasks as a sequence of single-instance problems, operating on separate (sometimes cropped and rescaled \cite{Chen22CVPR}) image regions.
Consequently, such methods need additional clicks if there are multiple similar foreground instances in an image, since each of those instances has to be processed separately with a disjoint set of user interactions, specifying the foreground and background. This is inefficient, since it is often the case that one object instance has to be considered as background for a different nearby instance, such that a refinement with a negative click becomes necessary for the current object of focus. 

In this work, we improve on both of the above issues by proposing a Dynamic Multi-object Interactive segmentation Transformer (\abbrev), a novel multi-instance approach for interactive segmentation that only requires a single forward pass through the feature extractor and that processes all relevant objects together, while learning a common background representation. Our approach is based on a novel Transformer-based iterative refinement architecture which determines instance level descriptors directly from the spatio-temporal click sequence. \abbrev\ dynamically generates queries to the Transformer that are conditioned on the backbone features at the click locations. These queries are updated during the refinement process whenever the network receives a new click, prompting the Transformer to output a new multi-instance segmentation. Thus, \abbrev\ removes the need to re-compute image-level features for each user interaction, while making more effective use of user clicks by handling multiple object instances together. 

The attention-based formulation of learning object representations from user interactions allows multiple objects to interact with each other and with the common background representations, thus enabling the network to estimate a better context from the input image. Fig.~\ref{fig:teaser} shows a typical example, highlighting \abbrev's capability to segment all the relevant objects in the input image using few clicks. An advantage of such a network formulation can be directly seen in the third refinement iteration, where a positive click on the unsegmented camel instance automatically removes the false positive region that was spilled over after segmenting a different camel instance nearby in the previous iteration. Existing approaches that perform sequential single-instance segmentation would have to first add a negative click to remove the false positive as part of the individual object refinement in the third iteration, thereby requiring additional annotation effort. 

In order to enable quantitative evaluation, 
we also propose a novel multi-instance interactive segmentation task (MIST) and a corresponding evaluation strategy. Compared to single-instance segmentation, MIST has the added complexity of requiring decisions which object to click on next, which is significantly harder than just deciding where to click next in a given single-instance error region. In particular, different users may apply different next-object selection strategies, and it is important that an interactive segmentation method is robust to this and always performs well. Hence, we propose to evaluate against a set of several different (but still basic) click sampling heuristics that are intended to span the expected variability of user types.

In summary, we propose \abbrev, a novel Transformer-based interactive segmentation method which uses a query bootstrapping mechanism to learn object representations from image-level features that are conditioned on the user interactions. We also model the iterative refinement process as temporal update steps for the queries to our Transformer module, which removes the need to re-compute image-level features. We evaluate \abbrev\ on the standard interactive segmentation benchmarks and show that it performs competitively in the single-instance setting, while outperforming existing state-of-the-art methods on multi-instance tasks.

\section{Related Work}
\label{sec:rel_work}

\PARbegin{Instance Segmentation.}  Methods that perform instance segmentation automatically generate masks for every object in the input image. Mask R-CNN~\cite{He17ICCV} is one of the most influential instance segmentation networks, which first generates object proposals and then segments these proposals using a mask head. Several other methods use a single-stage approach, either by grouping the pixels~\cite{Brabandere17CVPRW,Kong18CVPR,Neven19CVPR,Newell17NIPS,Novotny18ECCV}, or by employing dynamic networks~\cite{Tian20ECCV} on top of fully convolutional object detectors~\cite{Tian19ICCV}. After the success of Vision Transformers (ViT)~\cite{dosovitskiy21iclr} for image-level classification tasks, recent methods leverage Transformer-based architectures for performing instance segmentation.  MaskFormer~\cite{cheng2021maskformer} adds a mask classification head to DETR~\cite{Carion20ECCV} and models instance segmentation as a per-pixel classification task. Mask2Former~\cite{cheng2021mask2former} further extends MaskFormer by using a masked-attention Transformer decoder.
Unlike interactive segmentation methods, instance segmentation networks rely on segmenting a fixed set of classes and cannot incorporate user inputs for refinement.

\begin{figure*}

\definecolor{backbone_color}{HTML}{ff6d01}
\definecolor{feature_color}{RGB}{128, 128, 128}

\definecolor{click_sampling_color}{HTML}{9900ff}
\definecolor{transformer_color}{HTML}{34a853}

\definecolor{q1c}{HTML}{ebc728}
\definecolor{q2c}{HTML}{34a853}
\definecolor{q3c}{HTML}{ea4335}
\definecolor{qtc}{HTML}{4285f4}
\definecolor{qtp1c}{HTML}{4285f4}
\definecolor{bg1c}{HTML}{808080}
\definecolor{bg2c}{HTML}{bebebe}

\definecolor{c_red}{HTML}{ea4335}
\definecolor{c_blue}{HTML}{4285f4}
\definecolor{c_green}{HTML}{34a853}

\tikzstyle{querystyle}[black]=[anchor = center, rounded corners=1pt, rectangle, inner sep=2.5pt, fill=#1!70, draw=#1]

\begin{tikzpicture}[font=\footnotesize]

\node[inner sep=0pt] (input_im) {\includegraphics[width=30mm]{figures/architecture/input.jpg}};

\node[inner sep=0pt, above = 13mm of input_im.north, anchor=south, opacity=0.4] (initial_clicks) {\includegraphics[width=30mm]{figures/architecture/input.jpg}};
\draw [-latex, dashed] (input_im.130) -- (initial_clicks.230) node[pos=0.5, right] {Initial user clicks};

\node (backbone) [right = 6mm of input_im, anchor=south, trapezium, trapezium angle=82, minimum width=12.7mm, minimum height=14mm, rotate=-90, inner xsep=2.5pt, fill=backbone_color!20, draw=backbone_color!60, thick] {};
\node (backbone_text) [anchor=center,align=center]  at (backbone.center)  {Backbone};
\draw[-latex] (input_im.east) -- (backbone.south);

\node (decoder) [right = 3mm of backbone.north, anchor=north, trapezium, trapezium angle=82, minimum width=12.7mm, minimum height=14mm, rotate=90, inner xsep=2.5pt, fill=backbone_color!20, draw=backbone_color!60, thick] {};
\node (decoder_text) [anchor=center,align=center]  at (decoder.center)  {Feature \\Decoder};
\draw[-latex] (backbone.north) -- (decoder.north);

\node (features) [right  = 6mm of decoder.south, anchor=west, rectangle, minimum width=18mm, minimum height=9mm, fill=feature_color!20, draw=feature_color!60, thick, align=center] {Multi-Scale \\ Features $\mathcal{F}$};
\draw[-latex] (decoder.south) -- (features.west);

\node (feature_fusion) [right  = 6mm of features.east, anchor=west, rectangle, minimum width=12mm, minimum height=9mm, fill=backbone_color!20, draw=backbone_color!60, thick, align=center] {Feature \\ Fusion};
\draw[-latex] (features.east) -- (feature_fusion.west);

\node (fused_features) [right  = 6mm of feature_fusion.east, anchor=west, rectangle, minimum width=18mm, minimum height=9mm, fill=feature_color!20, draw=feature_color!60, thick, align=center] {Fused \\ Features};
\draw[-latex] (feature_fusion.east) -- (fused_features.west);

\node (click_sampler) [right = 23mm of initial_clicks.east, anchor=west, rectangle, minimum width=15mm, minimum height=19mm, fill=click_sampling_color!20, draw=click_sampling_color!60, thick, align=center] {Click \\ Feature \\ Sampling};

\node (qtplus1) [left = 3mm of click_sampler.135, anchor=east] {$c_{t+1}$};
\draw[-latex] (qtplus1.east) -- (click_sampler.135);

\node (qt) [below = 5mm of qtplus1.west, anchor=west] {$c_t$};
\draw[-latex]
  let
    \p1=(qt.east),
    \p2=(click_sampler.west)
  in
    (\x1,\y1) -- (\x2, \y1);
    
\node (q2) [below = 7mm of qt.west, anchor=west] {$c_2$};
\draw[-latex]
  let
    \p1=(q2.east),
    \p2=(click_sampler.west)
  in
    (\x1,\y1) -- (\x2, \y1);

\node (q1) [below = 3mm of q2.west, anchor=west] {$c_1$};
\draw[-latex]
  let
    \p1=(q1.east),
    \p2=(click_sampler.west)
  in
    (\x1,\y1) -- (\x2, \y1);

\draw[dotted, thick] (qt.south) -- (q2.north);

\draw[-latex,rounded corners]
  let
    \p1=(features.north),
    \p2=(click_sampler.south)
  in
    (\x1,\y1) -- ($(\x1,\y1)!0.4!(\x1,\y2)$) -| (\x2, \y2);

\node (c1) [anchor = center, circle,draw, inner sep=1pt, fill=q1c!50, draw=q1c] at ($(initial_clicks.center)+(-4.3mm, -3.7mm)$) {};
\draw [-latex, dashed, q1c] (c1.320) to [out=320,in=180] (q1.west);

\node (c2) [anchor = center, circle,draw, inner sep=1pt, fill=q2c!50, draw=q2c] at ($(initial_clicks.center)+(3mm, -3.4mm)$) {};
\draw [-latex, dashed, q2c] (c2.east) to [out=0,in=180] (q2.west);

\node (c3) [anchor = center, circle,draw, inner sep=1pt, fill=q3c!50, draw=q3c] at ($(initial_clicks.center)+(-5.7mm, 1.4mm)$) {};

\node (ct) [anchor = center, circle,draw, inner sep=1pt, fill=qtc!50, draw=qtc] at ($(initial_clicks.center)+(1.5mm, 6.7mm)$) {};
\draw [-latex, dashed, qtc] (ct.east) to [out=0,in=180] (qt.west);

\node (q1f) [anchor = center, right= 34mm of q1.west, rounded corners=1pt, rectangle, inner sep=2.5pt, fill=q1c!70, draw=q1c] {};
\node (q2f) [anchor = center, right= 34mm of q2.west, rounded corners=1pt, rectangle, inner sep=2.5pt, fill=q2c!70, draw=q2c] {};
\node (qtf) [anchor = center, right= 34mm of qt.west, rounded corners=1pt, rectangle, inner sep=2.5pt, fill=qtc!70, draw=qtc] {};
\draw[dotted, shorten <=4pt, shorten >=4pt, thick] (qtf.south) -- (q2f.north);
\draw[dotted, thick] (qt.south) -- (q2.north);

\node[inner sep=3pt,fit=(q1f)(q2f)(qtf)] (queries) {};
\scoped[on background layer]{
      \draw [draw=lightgray] (queries.south west) rectangle (queries.north east);}
\draw[-latex]
  let
    \p1=(click_sampler.east),
    \p2=(queries.west)
  in
    (\x1,\y2) -- (\x2, \y2);

\node (qtplus1f) [anchor = center, right= 34mm of qtplus1.west, rounded corners=1pt, rectangle, inner sep=2.5pt, fill=qtp1c!70, draw=qtp1c] {};

\node[inner sep=3pt,fit=(qtplus1f)] (queriesplus1) {};
\scoped[on background layer]{
      \draw [draw=lightgray] (queriesplus1.south west) rectangle (queriesplus1.north east);}
\draw[-latex]
  let
    \p1=(click_sampler.east),
    \p2=(queriesplus1.west)
  in
    (\x1,\y2) -- (\x2, \y2);

\node (qbg1) [anchor = center, below = 4.25mm of q1f.center, rounded corners=1pt, rectangle, inner sep=2.5pt, fill=bg1c!70, draw=bg1c] {};
\node (qbg2) [anchor = center, below = 2mm of qbg1.center, rounded corners=1pt, rectangle, inner sep=2.5pt, fill=bg2c!70, draw=bg2c] {};

\node[inner sep=3pt,fit=(qbg1)(qbg2)] (bgqueries) {};
\scoped[on background layer]{
      \draw [draw=lightgray] (bgqueries.south west) rectangle (bgqueries.north east);}
      
\node (bg_q_text) [inner sep=0pt, left = 2pt of bgqueries.west, anchor=east, align=center] {\scriptsize Learnd BG\\ Queries};

\node (tf_enc) [inner sep=0pt, draw=transformer_color!60, fill=transformer_color!20, rectangle, thick, fit= {($(queriesplus1.north east)+(5mm,0mm)$)($(queries.south east)+(28mm, -8mm)$)}] {};

\node (tf_enc_layer) [below right= 0.75mm and 0.75mm of tf_enc.north west, anchor=north west, rectangle, minimum width=15.2mm, minimum height=22.6mm, fill=transformer_color!40, draw=transformer_color!80] {};

\node (tf_enc_layer_text) [inner sep = 0pt, right = 1pt of tf_enc_layer.north east, anchor = north west] {$\times L_e$};

\node (tf_enc_ca) [below right = 0.75mm and 0.75mm of tf_enc_layer.north west, anchor=north east, rectangle, minimum width=21mm, rotate=90, fill=transformer_color!60, draw=transformer_color] {\scriptsize Masked Cross-Att.};
\node (tf_enc_sa) [right = 0.75mm of tf_enc_ca.south, anchor=north, rectangle, minimum width=21mm, rotate=90, fill=transformer_color!60, draw=transformer_color] {\scriptsize Self-Attention};
\node (tf_enc_ffn) [right = 0.75mm of tf_enc_sa.south, anchor=north, rectangle, minimum width=21mm, rotate=90, fill=transformer_color!60, draw=transformer_color] {\scriptsize FFN};

\node (tf_enc_text) [above= 1mm of tf_enc.south, anchor=south, inner sep = 0pt] {Encoder};

\draw[-latex,rounded corners]
  let
    \p1=(features.north),
    \p2=(click_sampler.south),
    \p3=(tf_enc_ca.west)
  in
    (\x1,\y1) -- ($(\x1,\y1)!0.4!(\x1,\y2)$) -| (\x3, \y3);

\draw[-latex]
  let
    \p1=(queries.east),
    \p2=(tf_enc_ca.north)
  in
    (\x1,\y1) -- (\x2, \y1);

\node (tf_dec) [inner sep=0pt, draw=transformer_color!60, fill=transformer_color!20, rectangle, fit= {($(tf_enc.north east)+(11mm,0mm)$)($(tf_enc.south east)+(30.2mm, 0mm)$)}] {};

\node (tf_dec_layer) [below right= 0.75mm and 0.75mm of tf_dec.north west, anchor=north west, rectangle, minimum width=10.4mm, minimum height=22.6mm, fill=transformer_color!40, draw=transformer_color!80] {};

\node (tf_dec_layer_text) [inner sep = 0pt, right = 1pt of tf_dec_layer.north east, anchor = north west] {$\times L_{d}$};

\node (tf_dec_ca) [below right = 0.75mm and 0.75mm of tf_dec_layer.north west, anchor=north east, rectangle, minimum width=21mm, rotate=90, fill=transformer_color!60, draw=transformer_color] {\scriptsize Cross Attention};
\node (tf_dec_ffn) [right = 0.75mm of tf_dec_ca.south, anchor=north, rectangle, minimum width=21mm, rotate=90, fill=transformer_color!60, draw=transformer_color] {\scriptsize FFN};

\node (tf_dec_text) [above right= 1mm and 2.5mm of tf_dec.south, anchor=south, inner sep = 0pt] {Decoder};

\draw[-latex,rounded corners]
  let
    \p1=(fused_features.north),
    \p2=(click_sampler.south),
    \p3=(tf_dec_ca.west)
  in
    (\x1,\y1) -- ($(\x1,\y1)!0.4!(\x1,\y2)$) -| (\x3, \y3);

\path
  let 
    \p1 =(tf_enc.east),
    \p2 =(tf_dec.west),
    \p3 = (qtf.center)
  in  
    node (qtfr) [querystyle=qtc] at ($(\x1,\y3)!0.5!(\x2,\y3)$) {};

\path
  let 
    \p1 =(tf_enc.east),
    \p2 =(tf_dec.west),
    \p3 = (q2f.center)
  in  
    node (q2fr) [querystyle=q2c] at ($(\x1,\y3)!0.5!(\x2,\y3)$) {};

\path
  let 
    \p1 =(tf_enc.east),
    \p2 =(tf_dec.west),
    \p3 = (q1f.center)
  in  
    node (q1fr) [querystyle=q1c] at ($(\x1,\y3)!0.5!(\x2,\y3)$) {};

\node (qtp1fr) [querystyle=qtp1c, above = 2mm of qtfr.center] {};
\node (qbg1fr) [querystyle=bg1c, below = 2mm of q1fr.center] {};
\node (qbg2fr) [querystyle=bg2c, below = 2mm of qbg1fr.center] {};
\draw[dotted, shorten <=4pt, shorten >=4pt, thick] (qtfr.south) -- (q2fr.north);

\node[inner sep=3pt,fit=(qtp1fr)(qbg2fr)] (refqueries) {};
\scoped[on background layer]{
      \draw [draw=lightgray] (refqueries.south west) rectangle (refqueries.north east);}

\draw[-latex]
  let
    \p1=(tf_enc.east),
    \p2=(refqueries.west),
    \p3=(queries.east)
  in
    (\x1,\y3) -- (\x2, \y3);

\draw[-latex]
  let
    \p1=(refqueries.east),
    \p2=(tf_dec_ca.north),
    \p3=(queries.east)
  in
    (\x1,\y3) -- (\x2, \y3);

\path
  let
    \p1=(tf_dec.east),
    \p2=(queries.east)
  in 
    node (output_im) [inner sep=0pt, anchor=west] at ($(\x1,\y2)+(7.8mm,0mm)$) {\includegraphics[width=30mm]{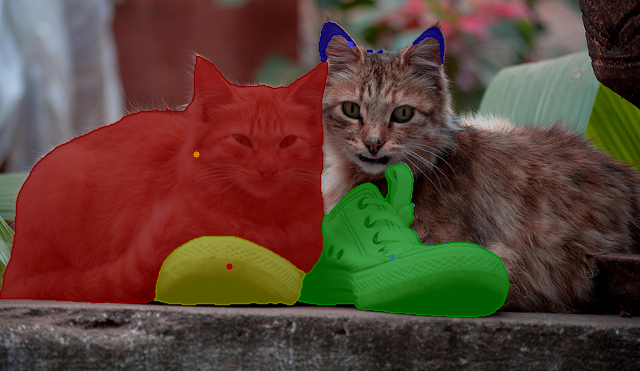}};

\draw[-latex]
  let
    \p1=(tf_dec.east),
    \p2=(output_im.west)
  in
    (\x1,\y2) -- (\x2, \y2) node [above, pos=0.5] {$\mathcal{M}^t$};

\node [anchor = center, circle,draw, inner sep=0.2pt, fill=q1c!50, draw=q1c] at ($(output_im.center)+(-4.3mm, -3.7mm)$) {};

\node [anchor = center, circle,draw, inner sep=0.2pt, fill=q2c!50, draw=q2c] at ($(output_im.center)+(3.3mm, -3.4mm)$) {};

\node [anchor = center, circle,draw, inner sep=0.2pt, fill=q3c!50, draw=q3c] at ($(output_im.center)+(-5.7mm, 1.4mm)$) {};

\node [anchor = center, circle,draw, inner sep=0.2pt, fill=qtc!50, draw=qtc] at ($(output_im.center)+(1.5mm, 6.7mm)$) {};

\node (cref) [anchor = center, circle,draw, inner sep=1pt, fill=qtp1c!50, draw=qtp1c] at ($(output_im.center)+(6.7mm, 1.4mm)$) {};

\node (refine) [below = 1mm of output_im.south, draw=c_blue, fill=c_blue!50, rounded corners, thick] {Refinement User Click};
\draw [-latex, densely dashed, c_blue!50, thick] (refine.north) to [out=90,in=270] (cref.south);


\draw[-latex,rounded corners, dashed, qtp1c]
  let
    \p1=(cref.north),
    \p2=(qtplus1.north)
  in
    (\x1,\y1) -- ($(\x1,\y1)+(0,17mm)$) -| (\x2, \y2);

\path
  let
    \p1=(cref.north),
    \p2=(qtplus1.north)
  in
    node (loop_text) [] at ($(\x1,\y1)!0.5!(\x2,\y1)+(0,14mm)$) {Iterative Refinement Loop ($\tau$ times)};

\path
  let
    \p1=(output_im.center),
    \p2=(fused_features.center)
  in 
    node (output_im_final) [inner sep=0pt, anchor=center] at (\x1,\y2) {\includegraphics[width=30mm]{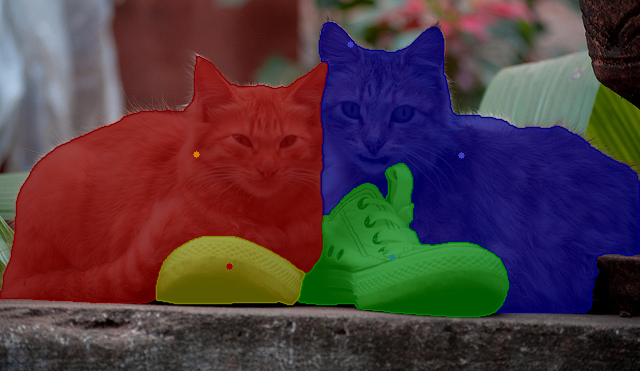}};

\node [anchor = center, circle,draw, inner sep=0.2pt, fill=q1c!50, draw=q1c] at ($(output_im_final.center)+(-4.3mm, -3.7mm)$) {};

\node [anchor = center, circle,draw, inner sep=0.2pt, fill=q2c!50, draw=q2c] at ($(output_im_final.center)+(3.3mm, -3.4mm)$) {};

\node [anchor = center, circle,draw, inner sep=0.2pt, fill=q3c!50, draw=q3c] at ($(output_im_final.center)+(-5.7mm, 1.4mm)$) {};

\node [anchor = center, circle,draw, inner sep=0.2pt, fill=qtc!50, draw=qtc] at ($(output_im_final.center)+(1.5mm, 6.7mm)$) {};

\node [anchor = center, circle,draw, inner sep=0.2pt, fill=qtp1c!50, draw=qtp1c] at ($(output_im_final.center)+(6.7mm, 1.4mm)$) {};

\draw[-latex,rounded corners]
  let
    \p1=(tf_dec.330),
    \p2=(click_sampler.south),
    \p3=(output_im_final.west)
  in
    (\x1,\y1) -- ($(\x1,\y1)!0.3!(\x3,\y1)$) |- (\x3, \y3) node [right, pos=0.25, inner sep=2pt] {$\mathcal{M}^{t+1}$};

\node[inner sep=8pt,fit=(backbone)(fused_features)] (run_once) {};
\scoped[on background layer]{
      \draw [draw=lightgray,pattern=north west lines, pattern color=lightgray] (run_once.south west) rectangle (run_once.north east);}
\node (run_once_text) [anchor=south east, above left= 0.5mm and 0.5mm of run_once.south east, fill=white, rounded corners, inner sep=1.5pt] {Executed once per image};

\end{tikzpicture}
  \caption{\abbrev\ consists of a backbone, a feature decoder, and an interactive Transformer. Point features at click locations at time $t$ are translated into queries which, along with the multi-scale features, are processed by a Transformer encoder-decoder structure to generate a set of output masks $\mathcal{M}^t$ for all the relevant objects. Based on $\mathcal{M}^t$, the user provides a new input click which is in turn used by the interactive Transformer to generate a new set of updated masks $\mathcal{M}^{t+1}$. This process is then iterated $\tau$ times until the masks are fully refined. }
  \label{fig:arch}
\end{figure*}

\PAR{Interactive Segmentation.} Earlier methods~\cite{boykov01cvpr, rother2004grabcut, Yu17ICIP} that perform interactive segmentation used graph-based optimisation techniques to translate user inputs to per-pixel segmentation masks. With the advent of deep learning, recent methods~\cite{Xu16CVPR, Castrejon17CVPR, Liew17ICCV, Lin16CVPR, mahadevan18bmvc, Kontogianni20ECCV, Sofiiukf20CVPR, Chen22CVPR,liu2022eccv} have been able to reduce the number of user interactions required for generating object masks. Most of these methods~\cite{mahadevan18bmvc, SofiiukArxiv21, Chen22CVPR} use positive and negative clicks to iteratively segment a foreground object by concatenating the input image with the click maps, along with the previous mask predictions, and then sending this combined representation through a deep network. This enables the network to learn the underlying representation of objects based on the input clicks. iADAPT~\cite{Kontogianni20ECCV} extends ITIS~\cite{mahadevan18bmvc} by considering user corrections as training examples during the testing phase, and updating the network parameters based on them, thereby aligning the training and testing domains. BRS~\cite{jang2019brs} is another interactive segmentation method that proposes an online update scheme for the network during testing, by constraining user click locations to have the corresponding click label. RITM~\cite{SofiiukArxiv21} improves the iterative training procedure introduced in ~\cite{mahadevan18bmvc}, and subsequently demonstrates better performance. The recent FocalClick~\cite{Chen22CVPR} approach builds upon the RITM~\cite{SofiiukArxiv21} pipeline, and uses a focus crop that is obtained based on user corrections during the refinement process. FocalClick achieves the state-of-the-art results on multiple instance segmentation datasets. 
Unlike \abbrev, all of these methods are designed to work with single instances, and also
need a complete forward pass for each refinement iteration.

Conceptually similar to our approach is that of Agustsson~\etal~\cite{Agustsson19CVPR}, who focus on full image segmentation. This is also a form of interactive multi-instance segmentation; however, every pixel in the image has to be assigned to a segment.
Their method is based on a two-stage Mask-RCNN, where the user specifies the object proposals with clicks on extreme object points, followed by scribbles for mask corrections.
Our method is more general, allowing the user to click on any object pixel and only requiring a minimum of one instead of four  clicks per object.

\section{Method}
\label{sec:method}
In a typical interactive segmentation process, the model first receives an input image along with the associated foreground (positive) clicks representing the objects that the user intends to segment. Based on this set of inputs, an interactive segmentation model predicts an initial set of segmentation masks corresponding to the clicks. These initial predictions are presented to the user so that they can provide a corrective click (which can be positive or negative) that is used to refine the previous network predictions. This process is repeated until the user receives a set of non-overlapping segmentation masks of satisfactory quality. 

Current state-of-the-art interactive segmentation models~\cite{mahadevan18bmvc, SofiiukArxiv21, Chen22CVPR, Kontogianni20ECCV, Sofiiukf20CVPR} perform this task sequentially, as their networks can handle only one foreground object at a time. These methods mostly use click maps, which are updated every time a user provides a new click and are then used to obtain a localized feature map from the feature extractor. \abbrev, on the other hand, can process multiple objects at once, and translates clicks to spatio-temporal data that is processed by an interactive transformer. This makes our model more efficient for mainly three reasons: (i) \abbrev\ just needs a single forward pass through the feature extractor to segment all the relevant foreground instances; (ii) the background is modeled jointly, and hence it reduces redundancy in negative clicks; and (iii) by annotating multiple objects jointly, these do not need to be repeatedly modeled as background for other foreground objects. 

\subsection{Network Architecture}
    
Following the state-of-the-art Transformer-based segmentation methods~\cite{zhu2020iclr,cheng2021maskformer, cheng2021mask2former, Athar22CVPR}, we use three basic components in our architecture: (i) a backbone network, (ii) a feature decoder, and (iii) a Transformer structure that processes the multi-scale image features from the feature decoder (Fig.~\ref{fig:arch}). Additionally, we also include a feature fusion module that fuses the multi-scale features to generate a fused feature map at the largest scale.
Our main contribution lies in the Transformer structure, which learns localized object descriptors directly from the user interactions without the need to pass them through the entire feature extractor. This results in an interactive segmentation network that is not only efficient in processing the user interactions, but also more practical since it can process multiple object instances at once. Since \abbrev\ can encode relationships between multiple objects in a given scene, it is naturally capable of segmenting multiple instances at once, which is a paradigm shift for interactive segmentation networks.

Fig.~\ref{fig:arch} shows the overall architecture of our network. It takes as input an RGB image $\mathcal{I} \in \mathbb{R}^{H \times W \times 3}$, and the corresponding set of user interactions $\mathcal{S}^t = \{c_1,c_2,...,c_t\}$ at timestep $t \in \{1,\dots, \mathcal{T}\}$, where $|\mathcal{T}|$ is the maximum number of refinement iterations for $\mathcal{I}$. Following the classic formulation of interactive segmentation that is used by existing works, we model the user interactions as positive and negative clicks, where positive clicks are placed on the foreground objects and the negative clicks on the background. 
Hence $\mathcal{S}^t = \{\mathcal{S}_{+}^t, \mathcal{S}_{-}^t\}$, where $\mathcal{S}_{+}^t = \{P_1^t, P_2^t, ..., P_n^t\}$ denote the positive clicks, and $\mathcal{S}_{-}^t = \{b_1^t, b_2^t,..., b_m^t\}$ denote the set of negative clicks  at time $t$. Here, $P_i^t \in S_{+}^t$ is a set of positive clicks that belong to object $o_i \in \mathcal{O}$. For existing interactive segmentation methods, $|\mathcal{O}|$ is always $1$ since they can process only one object at a time, which need not necessarily be the case for \abbrev{} as it is capable of handling multiple objects concurrently. All $P_i^t \in  S_{+}$, as well as $S_{-}$ are initialised as empty sets, and then updated when the user inputs a new click. The backbone processes $\mathcal{I}$ and extracts low-level features, which are then up-sampled by the feature decoder to produce feature maps $\mathcal{F} = \{f_{32}, f_{16}, f_8\}$ at multiple scales. These feature maps, along with the associated user interactions, up to time $t$, are then processed by the interactive Transformer.

\subsection{Interactive Transformer}
\label{sec:transformer}

The goal of our interactive Transformer is to generate segmentation masks $\mathcal{M}^t = \left\{M_1^t, M_2^t,..., M_n^t\right\}$ for all the relevant foreground objects at a given refinement timestep $t$, given the inputs $\mathcal{F}$ and the corresponding clicks $\mathcal{S}^t$.
These masks should be disjoint, \ie $M_i^t \cap M_j^t = \emptyset$ for all $i\neq j$.

\PAR{Dynamic Query Bootstrapping.} The queries used by the Transformer are dynamically generated using the input features $\mathcal{F}$ and the user clicks $\mathcal{S}^t$. 
To do this, we first sample the point features at every spatial location represented by each user click in $\mathcal{S}^t$ from all the feature scales in $\mathcal{F}$. Hence, if $Q^t$ denotes the set of queries at time $t$, then $q_j \in Q^t$ for click $c_j$ in $\mathcal{S}^t$ is generated as:
\begin{eqnarray}
    q_j = \frac{1}{|\mathcal{F}|}\sum_{f \in \mathcal{F}} f_{c_j}.
\end{eqnarray}

During the refinement process, the network receives a new interaction $c_{\mathit{t+1}}$ at the time step $\mathit{t+1}$, which is used to obtain an updated set of user clicks $\mathcal{S}^{t+1}$. To do this, if $c_{\mathit{t+1}}$ is a positive click then it is added to the corresponding object specific click set $P_j$ to obtain a new set of foreground clicks $\mathcal{S}_{+}^{t+1}$, else it is added to $\mathcal{S}_{-}$ to obtain $\mathcal{S}_{-}^{t+1}$. $\mathcal{S}^{t+1}$ is then used to obtain the updated queries $Q^{t+1}$, using the same process as explained above.
These queries are thus dynamically updated throughout the iterative process without the need to recompute $\mathcal{F}$, and the entire interactive segmentation process can work with multiple instances at once. 

In addition to the dynamic queries, we include a set of $K=9$ learnable queries for modeling the background without the use of any user guidance. These static background queries learn generic background representations and they reduce the background interactions that a user will have to perform. We also add a 3D positional encoding to $q_j$ where the first two dimensions represent the spatial location of the corresponding click in the image features and the third dimension represents the refinement timestep $t$.

\PAR{Instance Encoder.} The \abbrev\ instance encoder takes as input the queries $Q^t$ and the multi-scale feature maps $\mathcal{F}$. The encoder follows the structure of the masked attention Transformer decoder presented in~\cite{cheng2021mask2former} and leverages its capability of processing multi-scale features, which is important for dense pixel prediction tasks. Its main purpose is to enhance the initial click-based queries, such that they become more discriminative. We use $L_e=9$ layers, which are grouped into 3 blocks, each of which processes successive feature scales from $\mathcal{F}$. Every Transformer layer in the encoder consists of a masked attention module, a self-attention module and a feedforward network. Hence, our encoder block performs the following operations:%
\begin{align}
    Q_{l} &\xleftarrow[]{} \mathrm{MaskedCrossAttn}(Q_l, F, \mathcal{M}_{l-1}) + Q_{l-1},\\
    Q_{l} &\xleftarrow[]{} \mathrm{SelfAttn}(Q_{l}) + Q_{l},\\
    Q_{l} &\xleftarrow[]{} \mathrm{FFN}(Q_{l}) + Q_{l}.
\end{align}
Here, $Q_l$ represents the queries at $l^{\text{th}}$  layer; $\mathcal{M}_{l-1}$ is the attention mask produced from the binarized mask predictions from ${(l-1)}^{\text{th}}$ layer; $\mathrm{SelfAttn}$ is the multi-head self-attention module introduced in~\cite{vaswani17NIPS}; and $\mathrm{FFN}$ denotes a feedforward network. $\mathrm{MaskedCrossAttn}$ is a variant of cross-attention, where the attention operation is restricted to the foreground region represented by each query in the previous layer. Hence, each masked attention module performs the following operation:
\begin{equation}
    Q_{\mathit{l}} = \mathrm{softmax}(\mathcal{M}_{l-1} + Q_{l}K_{l})V_{l} + Q_{l-1},
\end{equation}
where $K_l$ and $V_l$ are the keys and values derived from $\mathcal{F}$ at the corresponding feature scale.

\PAR{Decoder.} While the goal of the encoder is to update the instance specific queries $Q$, the decoder updates the fused features $\mathcal{F^M}$. The fused features are obtained from the feature fusion module, which takes the multi-scale decoder features $\mathcal{F}$ as input and generates a fused feature map at the largest scale of $\mathcal{F}$. The feature fusion module consists of a convolutional layer, followed by an up-sampling layer with skip connections to upsample low resolution features from $\mathcal{F}$, which are then concatenated to generate $\mathcal{F^M}$. 

The decoder processes $\mathcal{F^M}$ using a set of $L_d=5$ Transformer layers. Each Transformer layer in our decoder consists of a cross-attention layer, followed by a feedforward network. Hence, each of the \abbrev\ decoder layers performs the following operations:
\begin{align}
    F_{l}^M &= \mathrm{softmax}(F_l^M K^T)V^T + F_{l-1}^M, \\
    F_{l}^M &= \mathrm{FFN}(F_{l}^M) + F_{\mathit{l}}^M.
\end{align}
Here $K$ and $V$ are again keys and values; however, they are obtained from the instance encoder's output $Q_{\mathit{out}}$ and not from $\mathcal{F^M}$. To get the final output masks, we take a dot product of the updated mask features $F_{out}^M$ with the click specific queries $\mathcal{Q_{\mathit{out}}}$, as done in~\cite{cheng2021mask2former}, and obtain a set of output mask probabilities. Since we have more than one query representing both the objects and the background, we use the per-pixel $\max$ operation over the corresponding set of queries to obtain instance specific masks for all the objects and the background. In the end, the discretized output masks are obtained by taking an $\mathrm{argmax}$ per pixel across the different instance predictions.

\section{Multi-instance Interactive Segmentation}
\label{sec:mist}
Existing interactive segmentation approaches address multi-instance segmentation as a sequence of single-instance tasks. \Ie, they pick one instance at a time, and then refine it either until the mask has a satisfactory quality, or until they have exhausted a specific click budget $\tau$ for that object. If there are multiple foreground objects in a single image, these methods generate overlapping object masks which have to be merged as an additional post-processing step in order to obtain the final masks. Also, since these objects are processed individually with disjoint click sets, some clicks can be redundant at an image-level. Hence, in this work we propose a novel multi-instance interactive segmentation task (MIST), where the goal of a user is to jointly annotate multiple object instances in the same input image.

Given an input image and a common set of user clicks, the MIST expects a corresponding method to generate non-overlapping instance masks for all relevant foreground objects. A major difference in this setting is that the background, and the corresponding negative clicks, are now common for all object instances. The MIST is a more challenging problem compared to the classical single-instance setting, since 
every refinement step can now lead to a positive click on any of the relevant objects or to a negative (background) click. Thus, extending an existing single-instance interactive segmentation method to the MIST is not trivial. 

\PAR{Automatic Evaluation.} It is also important to note that the user click patterns for the MIST may differ considerably between users. As a result, simulating the MIST for automatic evaluation is a challenge of its own.
In contrast to single-instance interactive segmentation benchmarks that have converged onto a deterministic next-click simulation strategy~\cite{mahadevan18bmvc,Kontogianni20ECCV,SofiiukArxiv21, Sofiiukf20CVPR,Chen22CVPR}, the refinement focus in the MIST may jump from one object to another in an arbitrary sequence, unless users are instructed to process the objects in an image according to a specific order. 
Since it is hard to predict what next-object/next-click selection strategies users will end up using in an actual interactive segmentation application, and since that choice will in turn depend on their impression of which strategies work best with the given segmentation method, it is not practical to assume a single, deterministic next-click simulation strategy. Instead, we postulate that a method that performs the MIST should ideally be robust against varying click patterns and next-object selection strategies. Hence, we propose a multi-fold evaluation based on three different next-object simulation patterns during refinement.

All of these click simulation strategies start by adding a single positive click to each of the foreground objects in that image to get an initial prediction. Based on this initial prediction, we choose an object $o_i$ according to one of the following strategies: (i) \textit{best}: choose the object that has the best IoU, compared to the ground truth mask; (ii) \textit{worst}: choose the object that has the worst IoU; and (iii) \textit{random}: choose a random object. In each of these strategies, only the objects that have not yet achieved the required segmentation quality will be sampled. Next, we place a simulated click $c_t$ on the largest error region of $o_i$. $c_t$ can now be (i) a positive click on $o_i$; (ii) a negative click on the background; or (iii) a positive click on another $o_{j}$. This process is repeated either until all the relevant objects are segmented, or until the image-level click budget $\tau$ is fully spent. 
We want to emphasize that we make no claim that those strategies (\textit{best}, \textit{worst}, \textit{random}) are close to optimal (in fact, we discuss several more effective strategies in the supplementary material). Instead, we intend for them to span the variability of possible next-object selection strategies to ensure that evaluated approaches generalize well to different users.

\PAR{Evaluation Metric.} The standard metric used by existing interactive segmentation benchmarks \cite{Rother04SIGGRAPH,mcguinness2010berkeley,Perazzi16CVPR,hariharan2011iccv} is the average number of clicks per object (NoC). Since the MIST is quite different from annotating instances individually, the NoC metric for interactive segmentation per object would not serve as a good evaluation metric.
Hence, we propose a new evaluation metric called Normalized Clicks per Image (NCI) for the multi-instance interactive segmentation task. NCI is an adaptation of NoC, where the number of clicks is now computed per image, instead of per-object, and is then normalized by the number of foreground objects in the image.
For NCI, we cap the number of clicks for an image based on the number of foreground objects. If an image has $|\mathcal{O}|$ foreground objects, then the total 
click budget for that image would be $\tau * |\mathcal{O}|$. Unlike the NoC metric, this cap is at an image level, and all of these clicks can be spent on a subset of objects if the corresponding algorithm so desires. Similar to the single-instance case, all objects that cannot be segmented to the desired quality level using this budget are marked as failure cases (counted as NFO), and the number of clicks for that image is set to the image-level click budget.
In addition, we also mark an image as a failed image (counted as NFI) if there is at least one object within that image that could not be segmented. 

\begin{table*}[t!]
\setlength{\tabcolsep}{1pt} %
\centering%
\footnotesize%
\begin{tabularx}{\textwidth}{p{2.7cm}cp{0.3cm}cp{0.3cm}YYp{0.1cm}YYp{0.1cm}YYp{0.1cm}YYp{0.1cm}YY}
\toprule%
&&&&&\multicolumn{2}{c}{GrabCut~\cite{Rother04SIGGRAPH}}&&\multicolumn{2}{c}{Berkeley~\cite{mcguinness2010berkeley}} && \multicolumn{2}{c}{SBD~\cite{hariharan2011iccv}} && \multicolumn{2}{c}{COCO MVal}&& \multicolumn{2}{c}{DAVIS~\cite{Perazzi16CVPR}}\\
\cmidrule{6-7}\cmidrule{9-10}\cmidrule{12-13}\cmidrule{15-16}\cmidrule{18-19}
Method & Backbone && Train Data && @85 $\downarrow$ & @90 $\downarrow$&& @85 $\downarrow$& @90 $\downarrow$&& @85 $\downarrow$& @90 $\downarrow$&& @85 $\downarrow$& @90  $\downarrow$&& @85 $\downarrow$& @90 $\downarrow$\\
\midrule
iFCN w/ GraphCut & - && PASCAL VOC && - &6.04&& - & 8.65 && - & - && - & - && - & -\\
ITIS~\cite{mahadevan18bmvc}&  DeepLabV3+ && SBD && - &5.6&& - & - && - & - && - & - && - & -\\
VOS-Wild~\cite{benard17arxiv} &  ResNet-101 && - && - &3.8&& - & - && - & - && - & - && - & -\\
iADAPT~\cite{Kontogianni20ECCV} & DeepLabV3+  && SBD && - &3.07&& - & 4.94 && - & - && - & - && - & -\\
\midrule
EdgeFlow~\cite{hao21iccvw} & hrnet18 && COCO+LVIS&& 1.60 & 1.72 & &-&2.40&&-&-&&-&-&& 4.54 & 5.77\\
RITM~\cite{SofiiukArxiv21} & hrnet32 && COCO+LVIS && 1.46& 1.56&& -& 2.10&&3.59 &5.71 & &-&-&& 4.11 &5.34\\
FocalClick~\cite{Chen22CVPR} & hrnet32 && COCO+LVIS&&1.64&1.80&& - & 2.36&&4.24& 6.51 &&-&-&& 4.01&5.39\\
f-BRS~\cite{Sofiiukf20CVPR}&hrnet32&&COCO+LVIS&&1.54&1.69&&1.64&2.44&&4.37&7.26&& 2.35 &3.44&&5.17&6.50\\
PseudoClick~\cite{liu2022eccv}&hrnet32&&COCO+LVIS&&-&\textcolor{red}{1.50}&&-&2.08&&-&\textbf{5.54}&&-&-&&\textcolor{red}{3.79}& \textcolor{red}{5.11}\\
\textbf{\abbrev} & hrnet32 &&COCO+LVIS&& 1.62 & 1.68 && \textcolor{red}{1.46} & \textcolor{red}{2.04}&& 3.83 & 6.35 && \textcolor{red}{2.35} & \textcolor{red}{3.14}&& 3.83 & 5.2\\
\midrule
FocalClick~\cite{Chen22CVPR}$^*$ & Resnet-50 && COCO+LVIS&&2.02&2.24&& 2.43 & 3.78&&5.10& 7.70 &&3.21&4.42&& 5.34&7.72.\\
\textbf{\abbrev} & Resnet-50 &&COCO+LVIS&& \textcolor{red}{1.68}& \textcolor{red}{1.82}&& \textcolor{red}{1.47}& \textcolor{red}{2.19}&& \textcolor{red}{3.93}& \textcolor{red}{6.56}&& \textcolor{red}{2.36}& \textcolor{red}{3.20}&& \textcolor{red}{4.10}& \textcolor{red}{5.45}\\
\midrule
FocalClick~\cite{Chen22CVPR} & Segformer-B0 && COCO+LVIS&&\textbf{1.40}&\textcolor{red}{1.66}&& \textcolor{red}{1.59} & 2.27&&4.56& 6.86 &&2.65&3.59&& 4.04&5.49.\\
\textbf{\abbrev} & Segformer-B0 &&COCO+LVIS&& 1.58& 1.68&& 1.61& \textcolor{red}{2.06}&& \textcolor{red}{3.89} & \textcolor{red}{6.48} && \textcolor{red}{2.47} & \textcolor{red}{3.28} && \textcolor{red}{3.85}& \textcolor{red}{5.08}\\
\midrule
\textbf{\abbrev} & Swin-T &&COCO+LVIS&& 1.64 & 1.78&& 1.39& 1.96&& 3.75 & 6.32 && 2.24& 3.14&& 3.87& 5.23\\
\textbf{\abbrev} & Swin-L &&COCO+LVIS&& 1.62 & 1.72&& \textbf{1.39}& 1.90&& \textbf{3.32} & 5.64 && \textbf{2.19}& \textbf{2.88}&& 3.80& 5.09\\
\midrule
FocalClick~\cite{Chen22CVPR} & Segformer-B3 && COCO+LVIS&&\textcolor{red}{1.44}&\textbf{1.50}&& 1.55 & 1.92&& 3.53 & \textcolor{red}{5.59} && 2.32 & 3.12 && \textbf{3.61} & \textbf{4.90}\\
saic-is\cite{faizov22icip} & Segformer-B4 && COCO+LVIS && 1.52 & 1.60 && \textcolor{red}{1.40} & \textbf{1.60} && \textcolor{red}{\textbf{3.44}} & 5.63 && - & - && 3.68 & 5.06  \\
\bottomrule
\end{tabularx}

\caption{NoC results on single-instance segmentation datasets grouped by the used backbone. Top results within a group are indicated in \textcolor{red}{red} and the overall top results in \textbf{bold}. Within groups we obtain state-of-the-art or competetive results.}
\label{tab:main}
\end{table*}

\section{Experiments}
\label{sec:experiments}

\PARbegin{Datasets and Metrics.} We evaluate \abbrev\ on an extensive range of datasets across two task settings. For the well established single-instance setting, we mainly use small-scale datasets such as GrabCut~\cite{Rother04SIGGRAPH}, Berkeley~\cite{mcguinness2010berkeley}, COCO MVal, and DAVIS~\cite{Perazzi16CVPR}. GrabCut and Berkeley are very small datasets with 50 and 96 images, respectively, mostly containing a single foreground object. COCO MVal is a subset of COCO~\cite{Lin2014ECCV} with a total of 800 images, and contains 10 objects from each object category. DAVIS~\cite{Perazzi16CVPR} is a video object segmentation dataset, which consists of 50 short videos for training and 20 for validation. Each video frame consists of a single salient foreground region, where object instances that belong together share a common mask. For evaluation, we use the subset of 345 randomly sampled images~\cite{jang2019brs} to be consistent with the existing interactive segmentation methods. Additionally, we also evaluate this task on SBD~\cite{hariharan2011iccv}, which is an extension of the PASCAL VOC~\cite{Everingham15IJCV} dataset with 10582 images containing 24125 object instances, with 6671 instances for validation. Although SBD contains multiple instances per image, it is adapted to the single-instance task setting by considering every image-instance pair as a separate data sample.

For evaluating the MIST, we use the large-scale instance segmentation dataset COCO~\cite{Lin2014ECCV} in conjunction with DAVIS17~\cite{PontTuset17Arxiv}, and SBD~\cite{hariharan2011iccv}. COCO is an image dataset with annotations for multiple image level tasks with 5k images for validation. DAVIS17 is an extension of the single-instance DAVIS~\cite{Perazzi16CVPR} dataset, which contains 30 validation videos with multiple segmented objects per-frame. In addition, we also use the annotations from LVIS~\cite{gupta2019lvis} for training \abbrev{}, where LVIS consists of a subset of COCO images with additional high-quality segmentation masks.

\PAR{Implementation Details.} For most experiments, we use a Swin Transformer~\cite{hu19iccv} as backbone, with a multi-scale deformable-attention Transformer~\cite{zhu2020iclr} on top to extract multi-scale features at 1/8, 1/16 and 1/32 scales. The encoder for our interactive Transformer follows the structure of the Transformer decoder in~\cite{cheng2021mask2former}. Specifically, there are 3 Transformer blocks in the encoder, each with 3 layers to process the feature maps at subsequent scales. For the interactive Transformer decoder, we use 5 layers of the cross-attention blocks as defined in Sec.~\ref{sec:transformer}.

The backbone is initialized with ImageNet~\cite{deng09cvpr} pretrained weights, while the feature decoder and the Transformer weights are initialized randomly. The entire network is trained end-to-end on the combined COCO+LVIS\cite{SofiiukArxiv21} dataset with an input resolution of 1024 $\times$ 1024 px for 50 epochs, and a batch size of 32 on 16 Nvidia A100 GPUs. We follow the iterative training strategy used in~\cite{SofiiukArxiv21}: we run a maximum of 3 iterative refinement steps to generate corrective clicks, based on the network output, for each object in an image during training.

\begin{table*}[t!]
\setlength{\tabcolsep}{1pt} %
\centering
\small
\begin{tabularx}{\textwidth}{p{1.8cm}cp{0.1cm}YcYYp{0.1cm}YYYYp{0.1cm}YYYY}
\toprule%
&&&\multicolumn{4}{c}{COCO}&&\multicolumn{4}{c}{SBD} && \multicolumn{4}{c}{DAVIS17}\\
\cmidrule{4-7}\cmidrule{9-12}\cmidrule{14-17}
Method & Backbone && NCI $\downarrow$ & NFO $\downarrow$& NFI $\downarrow$& IoU $\uparrow$ && NCI $\downarrow$& NFO $\downarrow$& NFI $\downarrow$ &IoU $\uparrow$ && NCI $\downarrow$& NFO $\downarrow$& NFI $\downarrow$& IoU $\uparrow$\\
\midrule
FocalClick~\cite{Chen22CVPR} & Segf-B0(best) && 7.31 & 19422& 3004 & 73.7 && 4.26& 1115& 599& 87.3 && 4.6& 802& 562& 84.6\\
FocalClick~\cite{Chen22CVPR} & Segf-B0(random) && 7.96 & 29240& 3463& 59.3&& 4.81& 2408& 838& 83.4 && 5.20& 1278& 685& 82.4\\
FocalClick~\cite{Chen22CVPR} & Segf-B0(worst) && 8.03 & 31234 & 3505 & 60.7 && 4.91 &  2723& 885 & 84.8 && 5.33 & 1433 & 689 & 81.6\\
\midrule
\abbrev & Segf-B0 (best) && 6.13 & 15219 & 2485 & 81.3 && 2.83 & 655 & 342 & 90.2 && 3.29 & 546 & 364 & 87.5\\
\abbrev & Segf-B0 (random)   && 6.04 & 12986 & 2431 & 84.9 && 2.76 & 528 & 313 & 90.6 && 3.27 & 549 & 356 & 87.9\\
\abbrev & Segf-B0 (worst)  && 6.02 & 19758 & 2414 & 83.0 && 2.75 &  841 & 315 & 90.3 && 3.25 & 707 & 354 & 87.1\\
\midrule
\abbrev & Swin-T(best)  && 6.07 & 14853 & 2460 & 81.8 && 2.75 & 624 & 327 & 90.3 && 3.20 & \textbf{501} & 348 & 87.7\\
\abbrev & Swin-T(random)  && 6.00 & \textbf{12710} & 2401 & \textbf{85.1} && 2.69 & \textbf{510} & 303 & \textbf{90.7} && \textbf{3.16} & 514 & \textbf{338} & \textbf{88.0} \\
\abbrev & Swin-T(worst) && \textbf{5.94} & 19309 & \textbf{2369} & 83.4 && \textbf{2.68} & 798 & \textbf{300} & 90.5 && 3.16 & 704 & 341 & 87.1\\

\bottomrule
\end{tabularx}
\caption{Results on the MIST using an IoU threshold of 85\%. NCI: normalised clicks per image, NFO: number of failed objects, NFI: number of failed images. All reported models are trained on COCO+LVIS.}
\label{tab:multi}
\end{table*}

\subsection{Comparison with the State-of-the-art}

\PARbegin{Single Instance Setting.} Although our model is designed to perform multi-instance interactive segmentation, we also apply it to the standard single-instance benchmark without any adaptations or re-training. In Tab~\ref{tab:main} we compare our results against previous methods, which are grouped based on the underlying network architecture and  the used training data. For this setting, we follow the same evaluation setting and the click sampling strategy adopted in previous works~\cite{mahadevan18bmvc, SofiiukArxiv21, Sofiiukf20CVPR, Chen22CVPR} and also set the click budget $\tau$ to 20.

Early deep learning models~\cite{mahadevan18bmvc, Kontogianni20ECCV, benard17arxiv} used larger backbones such as DeepLabV3+~\cite{Chen2018ECCV}, and were trained with small-scale image datasets such as PascalVOC~\cite{Everingham15IJCV}, while state-of-the-art interactive segmentation models mostly use HRNet~\cite{Wang19PAMI}.
To be consistent with these methods, we report the results for \abbrev{} using different commonly used backbone networks. Methods with comparable architectures are grouped together, and the corresponding best results within each group are marked in \textcolor{red}{red}. Although none of the \abbrev{} models were specifically trained to perform single-instance interactive segmentation, it outperforms comparable state-of-the-art networks for a majority of the datasets. Since vision transformers have recently emerged as a competitive alternative to CNNs, we additionally report \abbrev{} results with a Swin transformer~\cite{liu2021Swin}. 

\PAR{Multi-instance Interactive Segmentation (MIST).} For this experiment, we follow the MIST evaluation strategy described in Sec.~\ref{sec:mist} and use the proposed metrics (NCI, NFO, and NFI). In addition, we also report the average image-level IoU achieved after segmenting all the objects in an image. We use the validation sets of COCO~\cite{Lin2014ECCV}, SBD~\cite{hariharan2011iccv}, and DAVIS17~\cite{PontTuset17Arxiv} to evaluate our models and set $\tau=10$. 

As a baseline, we adapt FocalClick~\cite{Chen22CVPR} to the MIST setting. FocalClick is designed to work with a single object instance at a time, and processes objects sequentially to generate overlapping binary masks for each instance in an image. Hence, it cannot be directly used for automatic evaluation on the MIST, since the MIST click sampling strategy requires multi-instance segmentation masks to choose the object to refine in each iteration, and the MIST expects a non-overlapping instance map as final output. We fix these issues by adapting the evaluation pipeline of FocalClick to: (i) process all relevant objects sequentially using an initial click to obtain the initial predictions for all objects in an image; (ii) store both the intermediate IoUs and predictions at each refinement step, which are then used to choose the next object to refine and the corresponding simulated next click; and (iii) fuse the final predictions by performing an $argmax$ operation on the set of final predicted probabilities. 
We also tried to fuse the predictions at each intermediate refinement step but found it to perform worse. 
 
Tab.~\ref{tab:multi} shows the results of evaluating both FocalClick and \abbrev\ on the MIST using the three object sampling strategies (\textit{best}, \textit{worst}, and \textit{random}) explained in Sec.~\ref{sec:mist}. \abbrev\ outperforms FocalClick on all metrics and across all three datasets by a large margin. Additionally, the variance in performance across different sampling strategies is much smaller for \abbrev, demonstrating that it is more robust to variable user click patterns. \abbrev\ also generates segmentation masks of higher quality, as shown by the IoU values reported in Tab.~\ref{tab:multi}.

\subsection{Ablations}

Tab.~\ref{tab:abltation_main} reports the results of different ablations to analyze the impact of our network design choices (first group), and the positional encodings (second group) for \abbrev{}. For all of our ablation experiments, we use a Swin-T~\cite{He16CVPR} backbone with batch size 128, and evaluate it on the SBD~\cite{hariharan2011iccv} dataset for the MIST. Ablations on additional datasets are available in the supplementary.

\begin{table}[t!]
\setlength{\tabcolsep}{2pt} %
\centering
\small
\begin{tabularx}{\linewidth}{p{4.7cm}YYY}
\toprule
 & NCI$\downarrow$ & NFO $\downarrow$ & NFI$\downarrow$\\
\midrule
\abbrev{} (Swin-T) & \textbf{2.72} & \textbf{557} & \textbf{329}\\
- static background queries & 2.79 & 639 & 354\\
- Transformer decoder & 2.90 & 657 & 384\\
\midrule
- temporal positional encoding& 2.94 & 682 & 402\\
- spatial positional encoding& 2.90 & 671 & 395\\
- spatio-temporal positional encoding& 2.86 & 608 & 376\\
 
\bottomrule
\end{tabularx}

\caption{Ablation on the network design choices, always relative to the top line. All runs are repeated 3 times with random sampling and evaluated on SBD. All metrics use an IoU threshold of 85\%.}
\label{tab:abltation_main}
\end{table}

\begin{figure}
    \centering
    \begin{subfigure}[b]{\linewidth}
    \includegraphics[width=0.495\linewidth,trim={4cm 0cm 0.75cm 0cm},clip]{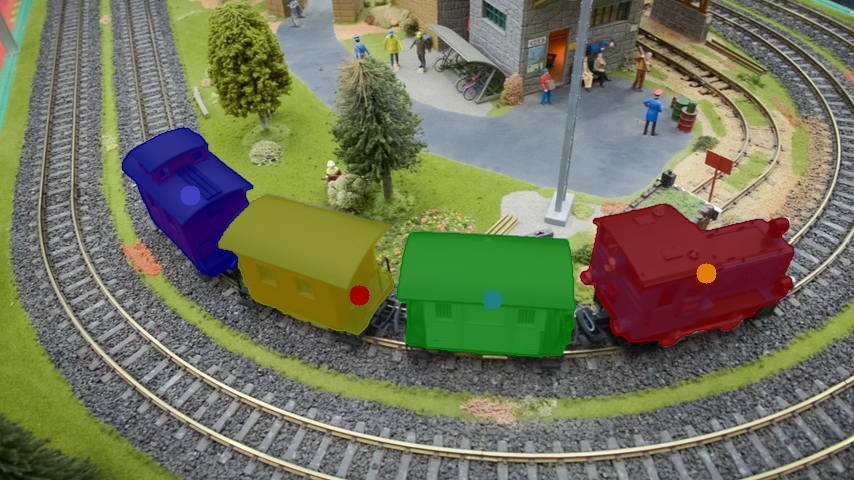}\hfill%
    \includegraphics[width=0.495\linewidth]{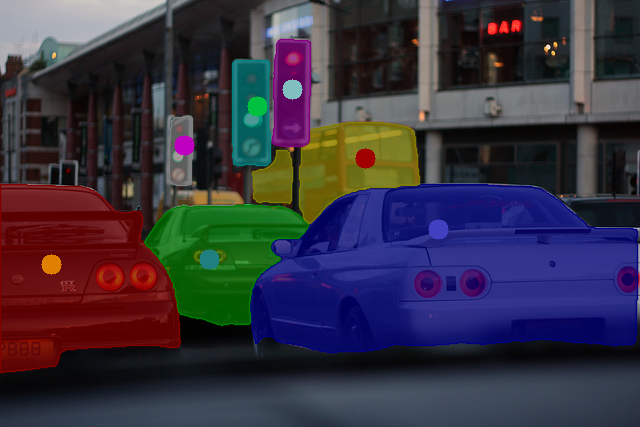}%
    \vspace{-3pt}
    \caption{Examples done after a single click per object.}
    \end{subfigure}
    
    \begin{subfigure}[b]{\linewidth}
    \includegraphics[width=0.495\linewidth]{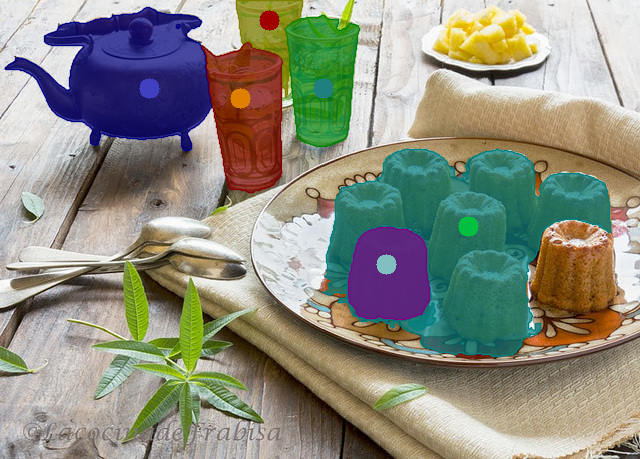}\hfill%
    \includegraphics[width=0.495\linewidth]{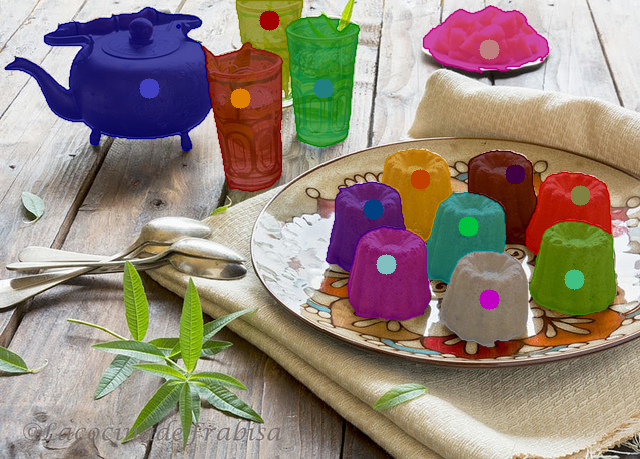}\hfill%

    \includegraphics[width=0.495\linewidth]{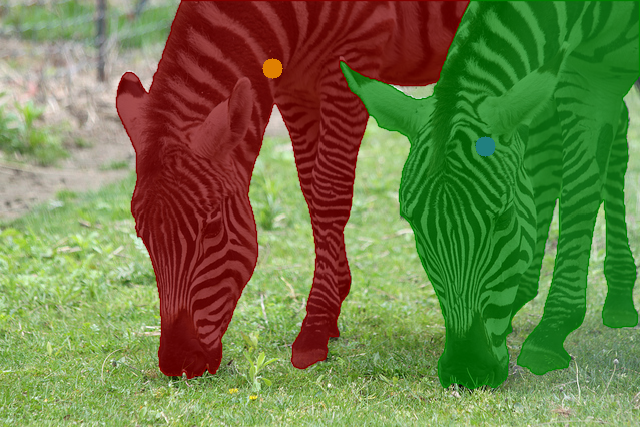}\hfill%
    \includegraphics[width=0.495\linewidth]{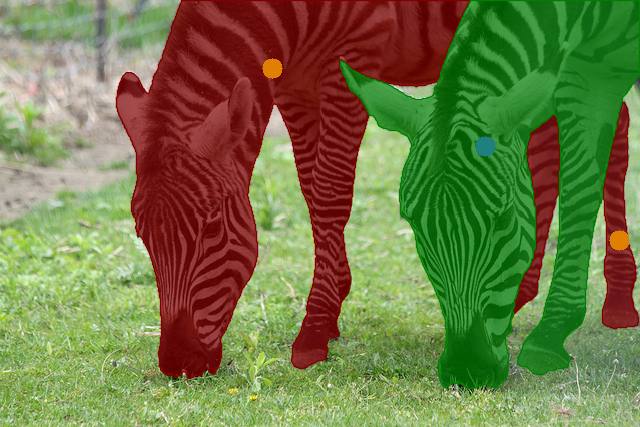}\hfill%
    \vspace{-3pt}
    \caption{Examples requiring refinement clicks.}
    \end{subfigure}
    \caption{Qualitative examples showing the annotation process with \abbrev{} for high-quality masks obtained with a single click per object and for cases that require additional refinements. Clicks are represented with colored dots.}    \label{fig:qualitative}
\end{figure}

\PAR{Transformer Decoder.} We ablate the effect of adding a Transformer decoder to the interactive transformer module. The decoder updates the fused image feature map at the highest resolution based on the instance queries. Discarding the Transformer decoder increases \abbrev's NCI from 2.72 to 2.90. It also adds an additional 100 failed objects, increasing the NFO from 557 to 657.

\PAR{Static Background Queries.} As mentioned in Sec.~\ref{sec:method}, we use a common set of $9$ learnable queries that model the background in an image. These queries learn generic background representations and help in reducing the number of background clicks required for performing interactive segmentation. As seen in Tab~\ref{tab:abltation_main}, adding the static background queries reduces the NCI from 2.79 to 2.72 and also the NFO from 639 to 557.

\PAR{Positional Encoding:} As clicks are interpreted as spatio-temporal data, we add a 3D positional encoding to the query features $Q$ and ablate its effect on the network performance on the MIST in the second section of Tab.~\ref{tab:abltation_main}. Removing the spatial and temporal positional encoding worsens the network performance by 0.18 and 0.22 NCI respectively. Not having any temporal encoding performs the worst with 2.94 NCI as compared to 2.72 for the full network. Temporal positional encodings seem to have a more significant impact compared to the spatial counterpart. This can be partly attributed to the fact that refinement clicks are often spatially close to each other, and hence the spatial positions alone do not provide good separation.

\subsection{Qualitative Results}
Fig.~\ref{fig:qualitative} shows several qualitative results produced by \abbrev{}.
The first row shows examples where a single click per object suffices to create well-defined segmentations for all objects.
The second and third row show examples where some refinement clicks are needed to arrive at the final masks.
While manually annotating images, one can notice that \abbrev{} mostly works with few clicks to create sharp masks and potential mistakes are often fixed with very few refinement clicks. Notice for example the single refinement click on one of the zebra's occluded legs in Fig.~\ref{fig:qualitative}(b) correctly fixed both legs.

\begin{figure}
    \centering
    \input{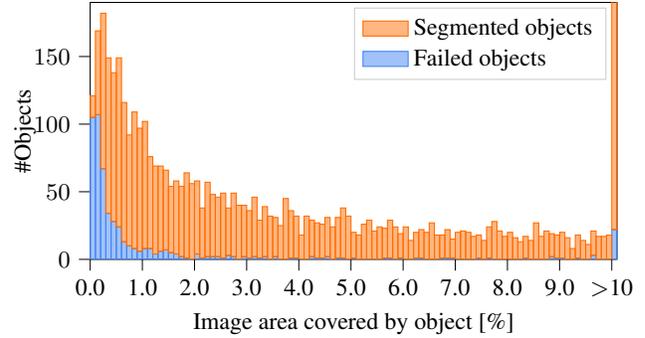}
    \vspace{-8pt}
    \caption{Failure cases analyzed by object size on the SBD dataset. The rightmost bin is truncated and contains 2582 segmented objects.}
    \label{fig:failure_hist}
\end{figure}

\subsection{Limitations}
\label{sec:limitations}

Fig.~\ref{fig:failure_hist} shows how the failure cases are distributed as a function of the relative area of an image they cover.
It can clearly be seen that the remaining failure cases of \abbrev\ mostly occur on objects covering a small image area.
One reason for this is that the highest resolution features map is downsampled by a factor of 4, making it harder to obtain very sharp masks.
Coarser object boundaries have a larger impact on the IoU for smaller objects.
Here, state-of-the-art single-instance segmentation approaches have the clear advantage that they process a zoomed-in crop around the object~\cite{Chen22CVPR,Sofiiukf20CVPR} and even additionally run a per-object mask refinement.
Such a high-resolution refinement step is orthogonal to our approach and could potentially be integrated into our pipeline, which we leave as future work.

\section{Conclusion}
\label{sec:conlusion}

We have introduced \abbrev, a novel Transformer-based interactive segmentation architecture that is capable of performing multi-instance segmentation, and a subsequent evaluation strategy. \abbrev{} dynamically generates instance queries based on the user clicks, and uses them within a Transformer architecture to generate and refine the corresponding instance segmentation masks. Unlike existing works, \abbrev{} can process user clicks for multiple instances at once without the need to re-compute image-level features. Our method achieves state-of-the-art results on multiple single-instance datasets and outperforms the FocalClick baseline on our novel MIST.

\PAR{Acknowledgements.}
This project was funded, in parts, by ERC Consolidator
Grant DeeVise (ERC-2017-COG-773161) and BMBF project NeuroSys-D (03ZU1106DA). Several experiments were performed using computing resources granted by RWTH Aachen University under project rwth1239, and  by the Gauss Centre for Supercomputing
e.V. through the John von Neumann Institute for Computing on the GCS Supercomputer JUWELS at J{\"u}lich Supercomputing Centre. We would like to thank Ali Athar, and Idil Esen Zulfikar for helpful discussions.

{\small
\bibliographystyle{ieee_fullname}
\bibliography{egbib}

\begin{thebibliography}{10}\itemsep=-1pt

\bibitem{Agustsson19CVPR}
Eirikur Agustsson, Jasper~RR Uijlings, and Vittorio Ferrari.
\newblock {Interactive Full Image Segmentation by Considering All Regions
  Jointly}.
\newblock In {\em CVPR}, 2019.

\bibitem{Athar22CVPR}
Ali Athar, Jonathon Luiten, Alexander Hermans, Deva Ramanan, and Bastian Leibe.
\newblock {HODOR: High-level Object Descriptors for Object Re-segmentation in
  Video Learned from Static Images}.
\newblock In {\em CVPR}, 2022.

\bibitem{benard17arxiv}
Arnaud Benard and Michael Gygli.
\newblock Interactive video object segmentation in the wild.
\newblock In {\em arXiv preprint arXiv:1801.00269}, 2017.

\bibitem{boykov01cvpr}
Yuri Boykov and Marie-pierre Jolly.
\newblock Interactive graph cuts for optimal boundary \& region segmentation of
  objects in n-d images.
\newblock In {\em ICCV}, 2001.

\bibitem{Carion20ECCV}
Nicolas Carion, Francisco Massa, Gabriel Synnaeve, Nicolas Usunier, Alexander
  Kirillov, and Sergey Zagoruyko.
\newblock End-to-end object detection with transformers.
\newblock In {\em ECCV}, 2020.

\bibitem{Castrejon17CVPR}
Lluis Castrejon, Kaustav Kundu, Raquel Urtasun, and Sanja Fidler.
\newblock Annotating object instances with a polygon-rnn.
\newblock In {\em CVPR}, 2017.

\bibitem{Chen2018ECCV}
Liang-Chieh Chen, Yukun Zhu, George Papandreou, Florian Schroff, and Hartwig
  Adam.
\newblock Encoder-decoder with atrous separable convolution for semantic image
  segmentation.
\newblock In {\em ECCV}, 2018.

\bibitem{Chen22CVPR}
Xi Chen, Zhiyan Zhao, Yilei Zhang, Manni Duan, Donglian Qi, and Hengshuang
  Zhao.
\newblock Focalclick: Towards practical interactive image segmentation.
\newblock In {\em CVPR}, 2022.

\bibitem{cheng2021mask2former}
Bowen Cheng, Ishan Misra, Alexander~G. Schwing, Alexander Kirillov, and Rohit
  Girdhar.
\newblock Masked-attention mask transformer for universal image segmentation.
\newblock In {\em CVPR}, 2022.

\bibitem{cheng2021maskformer}
Bowen Cheng, Alexander~G. Schwing, and Alexander Kirillov.
\newblock Per-pixel classification is not all you need for semantic
  segmentation.
\newblock In {\em NeurIPS}, 2021.

\bibitem{Brabandere17CVPRW}
Bert De~Brabandere, Davy Neven, and Luc Van~Gool.
\newblock Semantic instance segmentation for autonomous driving.
\newblock In {\em CVPR Workshops}, 2017.

\bibitem{deng09cvpr}
Jia Deng, Wei Dong, Richard Socher, Li-Jia Li, Kai Li, and Li Fei-Fei.
\newblock Imagenet: A large-scale hierarchical image database.
\newblock In {\em CVPR}, 2009.

\bibitem{dosovitskiy21iclr}
Alexey Dosovitskiy, Lucas Beyer, Alexander Kolesnikov, Dirk Weissenborn,
  Xiaohua Zhai, Thomas Unterthiner, Mostafa Dehghani, Matthias Minderer, Georg
  Heigold, Sylvain Gelly, Jakob Uszkoreit, and Neil Houlsby.
\newblock An image is worth 16x16 words: Transformers for image recognition at
  scale.
\newblock In {\em ICLR}, 2021.

\bibitem{Everingham15IJCV}
M. Everingham, S.~M.~A. Eslami, L. Van~Gool, C.~K.~I. Williams, J. Winn, and A.
  Zisserman.
\newblock The pascal visual object classes challenge: A retrospective.
\newblock In {\em IJCV}, 2015.

\bibitem{faizov22icip}
Boris Faizov, Vlad Shakhuro, and Anton Konushin.
\newblock Interactive image segmentation with transformers.
\newblock In {\em ICIP}, 2022.

\bibitem{gupta2019lvis}
Agrim Gupta, Piotr Dollar, and Ross Girshick.
\newblock {LVIS}: A dataset for large vocabulary instance segmentation.
\newblock In {\em CVPR}, 2019.

\bibitem{hao21iccvw}
Yuying Hao, Yi Liu, Zewu Wu, Lin Han, Yizhou Chen, Guowei Chen, Lutao Chu,
  Shiyu Tang, Zhiliang Yu, Zeyu Chen, et~al.
\newblock Edgeflow: Achieving practical interactive segmentation with
  edge-guided flow.
\newblock {\em ICCVW}, 2021.

\bibitem{hariharan2011iccv}
Bharath Hariharan, Pablo Arbel{\'a}ez, Lubomir Bourdev, Subhransu Maji, and
  Jitendra Malik.
\newblock Semantic contours from inverse detectors.
\newblock In {\em ICCV}, 2011.

\bibitem{He17ICCV}
K. He, G. Gkioxari, P. Doll{\'a}r, and R. Girshick.
\newblock Mask {R-CNN}.
\newblock In {\em ICCV}, 2017.

\bibitem{He16CVPR}
Kaiming He, Xiangyu Zhang, Shaoqing Ren, and Jian Sun.
\newblock Deep residual learning for image recognition.
\newblock In {\em CVPR}, 2016.

\bibitem{hu19iccv}
Han Hu, Zheng Zhang, Zhenda Xie, and Stephen Lin.
\newblock Local relation networks for image recognition.
\newblock In {\em ICCV}, 2019.

\bibitem{jang2019brs}
Won-Dong Jang and Chang-Su Kim.
\newblock Interactive image segmentation via backpropagating refinement scheme.
\newblock In {\em CVPR}, 2019.

\bibitem{Kong18CVPR}
Shu Kong and Charless~C Fowlkes.
\newblock Recurrent pixel embedding for instance grouping.
\newblock In {\em CVPR}, 2018.

\bibitem{Kontogianni20ECCV}
Theodora Kontogianni, Michael Gygli, Jasper Uijlings, and Vittorio Ferrari.
\newblock Continuous adaptation for interactive object segmentation by learning
  from corrections.
\newblock In {\em ECCV}, 2020.

\bibitem{Liew17ICCV}
JunHao Liew, Yunchao Wei, Wei Xiong, Sim-Heng Ong, and Jiashi Feng.
\newblock Regional interactive image segmentation networks.
\newblock In {\em ICCV}, 2017.

\bibitem{Lin16CVPR}
Di Lin, Jifeng Dai, Jiaya Jia, Kaiming He, and Jian Sun.
\newblock Scribblesup: Scribble-supervised convolutional networks for semantic
  segmentation.
\newblock In {\em CVPR}, 2016.

\bibitem{Lin2014ECCV}
Tsung-Yi Lin, Michael Maire, Serge~J. Belongie, James Hays, Pietro Perona, Deva
  Ramanan, Piotr Doll{\'a}r, and C.~Lawrence Zitnick.
\newblock Microsoft coco: Common objects in context.
\newblock In {\em ECCV}, 2014.

\bibitem{liu2022eccv}
Qin Liu, Meng Zheng, Benjamin Planche, Srikrishna Karanam, Terrence Chen, Marc
  Niethammer, and Ziyan Wu.
\newblock Pseudoclick: Interactive image segmentation with click imitation.
\newblock In {\em ECCV}, 2022.

\bibitem{liu2021Swin}
Ze Liu, Yutong Lin, Yue Cao, Han Hu, Yixuan Wei, Zheng Zhang, Stephen Lin, and
  Baining Guo.
\newblock Swin transformer: Hierarchical vision transformer using shifted
  windows.
\newblock In {\em ICCV}, 2021.

\bibitem{lohchilov19iclr}
Ilya Loshchilov and Frank Hutter.
\newblock Decoupled weight decay regularization.
\newblock In {\em ICLR}, 2019.

\bibitem{mahadevan18bmvc}
Sabarinath Mahadevan, Paul Voigtlaender, and Bastian Leibe.
\newblock Iteratively trained interactive segmentation.
\newblock In {\em British Machine Vision Conference (BMVC)}, 2018.

\bibitem{mcguinness2010berkeley}
Kevin McGuinness and Noel~E O’connor.
\newblock A comparative evaluation of interactive segmentation algorithms.
\newblock {\em Pattern Recognition}, 2010.

\bibitem{milletari163dv}
Fausto Milletari, Nassir Navab, and Seyed-Ahmad Ahmadi.
\newblock V-net: Fully convolutional neural networks for volumetric medical
  image segmentation.
\newblock In {\em 3DV}, 2016.

\bibitem{Neven19CVPR}
Davy Neven, Bert~De Brabandere, Marc Proesmans, and Luc~Van Gool.
\newblock Instance segmentation by jointly optimizing spatial embeddings and
  clustering bandwidth.
\newblock In {\em CVPR}, 2019.

\bibitem{Newell17NIPS}
Alejandro Newell, Zhiao Huang, and Jia Deng.
\newblock Associative embedding: End-to-end learning for joint detection and
  grouping.
\newblock In {\em NeurIPS}, 2017.

\bibitem{Novotny18ECCV}
D. Novotny, S. Albanie, D. Larlus, and A. Vedaldi.
\newblock Semi-convolutional operators for instance segmentation.
\newblock In {\em ECCV}, 2018.

\bibitem{Perazzi16CVPR}
Federico Perazzi, Jordi Pont-Tuset, Brian McWilliams, Luc~Van Gool, Markus
  Gross, and Alexander Sorkine-Hornung.
\newblock A benchmark dataset and evaluation methodology for video object
  segmentation.
\newblock In {\em CVPR}, 2016.

\bibitem{PontTuset17Arxiv}
Jordi Pont-Tuset, Federico Perazzi, Sergi Caelles, Pablo Arbel\'aez, Alexander
  Sorkine-Hornung, and Luc {Van Gool}.
\newblock The 2017 davis challenge on video object segmentation.
\newblock {\em arXiv}, 2017.

\bibitem{rother2004grabcut}
Carsten Rother, Vladimir Kolmogorov, and Andrew Blake.
\newblock " grabcut" interactive foreground extraction using iterated graph
  cuts.
\newblock {\em TOG}, 2004.

\bibitem{Rother04SIGGRAPH}
Carsten Rother, Vladimir Kolmogorov, and Andrew Blake.
\newblock "grabcut": Interactive foreground extraction using iterated graph
  cuts.
\newblock In {\em SIGGRAPH}, 2004.

\bibitem{Sofiiukf20CVPR}
Konstantin Sofiiuk, Ilia Petrov, Olga Barinova, and Anton Konushin.
\newblock f-brs: Rethinking backpropagating refinement for interactive
  segmentation.
\newblock In {\em CVPR}, 2020.

\bibitem{SofiiukArxiv21}
Konstantin Sofiiuk, Ilia Petrov, and Anton Konushin.
\newblock Reviving iterative training with mask guidance for interactive
  segmentation.
\newblock {\em arXiv preprint arXiv:2102.06583}, 2021.

\bibitem{Tian20ECCV}
Zhi Tian, Chunhua Shen, and Hao Chen.
\newblock Conditional convolutions for instance segmentation.
\newblock In {\em ECCV}, 2020.

\bibitem{Tian19ICCV}
Zhi Tian, Chunhua Shen, Hao Chen, and Tong He.
\newblock {FCOS}: Fully convolutional one-stage object detection.
\newblock In {\em Proc. Int. Conf. Computer Vision (ICCV)}, 2019.

\bibitem{vaswani17NIPS}
Ashish Vaswani, Noam Shazeer, Niki Parmar, Jakob Uszkoreit, Llion Jones,
  Aidan~N. Gomez, \L{}ukasz Kaiser, and Illia Polosukhin.
\newblock Attention is all you need.
\newblock In {\em NeurIPS}, NIPS'17, 2017.

\bibitem{Wang19PAMI}
Jingdong Wang, Ke Sun, Tianheng Cheng, Borui Jiang, Chaorui Deng, Yang Zhao,
  Dong Liu, Yadong Mu, Mingkui Tan, Xinggang Wang, Wenyu Liu, and Bin Xiao.
\newblock Deep high-resolution representation learning for visual recognition.
\newblock {\em IEEE TPAMI}, 2019.

\bibitem{Xu16CVPR}
Ning Xu, Brian Price, Scott Cohen, Jimei Yang, and Thomas~S Huang.
\newblock Deep interactive object selection.
\newblock In {\em CVPR}, 2016.

\bibitem{Yu17ICIP}
Hongkai Yu, Youjie Zhou, Hui Qian, Min Xian, Yuewei Lin, Dazhou Guo, Kang
  Zheng, Kareem Abdelfatah, and Song Wang.
\newblock Loosecut: Interactive image segmentation with loosely bounded boxes.
\newblock In {\em ICIP}, 2017.

\bibitem{zhu2020iclr}
Xizhou Zhu, Weijie Su, Lewei Lu, Bin Li, Xiaogang Wang, and Jifeng Dai.
\newblock Deformable detr: Deformable transformers for end-to-end object
  detection.
\newblock {\em ICLR}, 2020.

\end{thebibliography}
}
\clearpage

\appendix
\twocolumn[
\begin{center}
  {\Large \bf \Large{\abbrev~: Supplementary Material } \par}
  \vspace*{12pt}
\end{center}
]

\newcommand{\secMethod}{Sec.~\textcolor{red}{3}\@\xspace}
\newcommand{\secExperiments}{Sec.~\textcolor{red}{5}\@\xspace}
\newcommand{\secAblations}{Sec.~\textcolor{red}{5.2}\@\xspace}
\newcommand{\secMIST}{Sec.~\textcolor{red}{4}\@\xspace}

\renewcommand\thesection{\Roman{section}}
\renewcommand{\thetable}{\Roman{table}}

\begin{abstract}
In this supplementary material, we provide some additional details, ablations and also qualitative results for our approach.
\end{abstract}

\section{Additional Implementation Details}

As explained in \secMethod, \abbrev{} takes an image as input, and generates a set of output masks  probabilities $Y^t = \{Y_{1}^t, Y_{2}^t,..., Y_{n}^t\}$ by multiplying the instance encoder's output $Q_{out}^t$ with the output feature map $F_{out}^M$ at timestep $t$. Here, each $Y_{i}$ represents a set of object probabilities for $o_i \in \{\mathcal{O}, bg\}$, where $bg$ represents the background. The final segmentation masks $\mathcal{M}^t$ are then obtained by first taking a $\max$ per pixel over each $Y_i$, and then an $\mathrm{argmax}$ over the entire $Y^t$.

\PAR{Training.}During training, we apply a weighted sum of the binary cross-entropy loss and the dice loss $L = \lambda_{1}L_{bce} + \lambda_{2}L_{dice}$~\cite{milletari163dv} on the individual mask probabilities. The network is trained end-to-end using the AdamW~\cite{lohchilov19iclr} optimizer for 50 epochs with a batch size of 32 and an initial learning rate of $1e-4$, which is then decayed by $0.1$ after $44$ and $48$ epochs respectively. The models used for ablation are trained with batch size 128 and an initial learning rate of $5e-4$.

\section{MIST: Additional Evaluation Strategies}

In \secMIST\, we discussed a number of click simulation strategies that could potentially capture some of the user patterns for the MIST. Since these simulation strategies are not exhaustive, we discuss a few more such next-click strategies that could be used to better emulate how a user might perform a MIST. We also evaluate \abbrev{} on all of these strategies in Tab.~\ref{tab:new_strat}, and once again confirm that our model is robust against different user patterns.

\PAR{Round-robin:} The round-robin strategy assigns a click window of $\beta$ clicks for each of the objects in an image. Here, an object is chosen randomly and after the current object of focus exhausts all the $\beta$ clicks, the next random object is chosen and then refined until completion. Once all the objects in the input image are processed in this manner, the round-robin strategy revisits all the failed objects and then tries to refine their segmentation masks either until all the objects are fully segmented, or until the image-level click budget $\tau * |\mathcal{O}|$ is fully used up.

\PAR{Worst with limit:} Here, in each iteration we choose the object with the worst IoU, as we also do in the \textit{worst} strategy described in \secMIST, but we additionally add a per-object click limit $\beta$ to each object. Upon selecting the next worst object, we first check if this object has not reached its click limit and if it did, we skip this object until all objects have either been segmented or reached their limit.
After this is the case, we switch to the \textit{best} strategy and try to segment the remaining objects as usual until the image budget is used up or all objects are segmented.
The intuition behind this strategy is that a user will try to improve the biggest errors first, but they will notice when an object is not segmentable by the method at hand and rather spend more clicks on objects which can be segmented properly.

\PAR{Max-distance:} In this strategy, we again start by adding a positive click to each of the foreground objects. During refinement, the next click is simply sampled on the pixel with the maximum distance from the distance transform computed on the error region of the entire semantic map that includes the segmentation masks for all objects in an image. If the chosen pixel falls on an object, then a corresponding positive click is added to that object, and if it doesn't, then it is classified as a negative click.

For the results reported in Tab.~\ref{tab:new_strat}, we use $\tau=10$ and $\beta=10$. All of the strategies work and \textit{worst with limit} actually results in a lower number of failed objects in all cases, while having comparable NCI. The \textit{max-distance} strategy is actually amongst the worst, resulting in the highest number of failed images. A potential reason could be that due to the joint maximum distance transform over all object errors, the clicks are no longer sampled in order to specifically correct a mistake with respect to one object and are thus less targeted. This in turn might lead to failed objects, where the other strategies that rely on a per-object distance transform actually are able to sample clicks in more useful locations.

\begin{table*}[t!]
\setlength{\tabcolsep}{1pt} %
\centering
%
\small
\begin{tabularx}{\textwidth}{p{1.8cm}cp{0.1cm}YcYYp{0.1cm}YYYYp{0.1cm}YYYY}
\toprule%

&&&\multicolumn{4}{c}{COCO}&&\multicolumn{4}{c}{SBD} && \multicolumn{4}{c}{DAVIS17}\\
\cmidrule{4-7}\cmidrule{9-12}\cmidrule{14-17}

Backbone & Strategy && NCI $\downarrow$ & NFO $\downarrow$& NFI $\downarrow$& IoU $\uparrow$ && NCI $\downarrow$& NFO $\downarrow$& NFI $\downarrow$ &IoU $\uparrow$ && NCI $\downarrow$& NFO $\downarrow$& NFI $\downarrow$& IoU $\uparrow$\\

\midrule
Resnet50 & best  && 6.20 & 15690 & 2508 & 80.9 && 2.87 & 677 & 352 & 90.0 && 3.42 & 572 & 380 & 86.9\\
Resnet50 & random  && 6.13 & 13554 & 2461 & 84.4 && 2.81 & 559 & 329 & 90.4 && 3.39 & 580 & 375 & 87.3\\
Resnet50 & worst  && 6.09 & 20224 & 2447 & 82.6 && 2.78 & 870 & 324 & 90.2 && 3.36 & 773 & 375 & 86.2\\
Resnet50 & max-distance  && 6.82 & 14786 & 2890 & 85.2 && 3.24 & 762 & 471 & 90.7 && 3.57 & 627 & 405 & 87.5\\
Resnet50 & round-robin  && 6.51 & 15534 & 2501 & 83.4 && 3.50 & 620 & 335 & 90.3 && 4.07 & 609 & 373 & 87.1\\
Resnet50 & worst with limit  && 6.09 & 13249 & 2444 & 84.1 && 2.79 & 541 & 326 & 90.4 && 3.36 & 570 & 375 & 87.2\\
\hdashline
Resnet50 & mean  && 6.31 & 15506 & 2542 & 83.4 && 3.00 & 671 & 356 & 90.3 && 3.53 & 622 & 380 & 87.0\\
Resnet50 & std  && 0.30 & 2519 & 173 & 1.5 && 0.30 & 126 & 57 & 0.23 && 0.28 & 77 & 12 & 0.45\\

\midrule
Segf-B0 & best  && 6.13 & 15219 & 2485 & 81.3 && 2.83 & 655 & 342 & 90.2 && 3.29 & 546 & 364 & 87.5\\
Segf-B0 & random  && 6.04 & 12986 & 2431 & 84.9 && 2.76 & 528 & 313 & 90.6 && 3.27 & 549 & 356 & 87.9\\
Segf-B0 & worst  && 6.02 & 19758 & 2414 & 83.0 && 2.75 &  841 & 315 & 90.3 && 3.25 & 707 & 354 & 87.1\\
Segf-B0 & max-distance  && 6.79 & 14588 & 2885 & 85.5 && 3.18 & 735 & 441 & 90.8 && 3.42 & 592 & 388 & 88.2\\
Segf-B0 & round-robin  && 6.42 & 14608 & 2452 & 84.0 && 3.47 & 609 & 339 & 90.5 && 3.95 & 573 & 362 & 87.8\\
Segf-B0 & worst with limit  && 6.03 & 12745 & 2425 & 84.6 && 2.75 & 519 & 320 & 90.5 && 3.25 & 526 & 354 & 87.9\\
\hdashline
Segf-B0 & mean  && 6.24 & 14984 & 2515 & 83.8 && 2.96 & 648 & 345 & 90.5 && 3.40 & 582 & 363 & 87.7\\
Segf-B0 & std  && 0.31 & 2536 & 183 & 1.5 && 0.30 & 124 & 48 & 0.21 && 0.27 & 65 & 13 & 0.38\\

\midrule
hrnet32 & best  && 6.14 & 15092 & 2506 & 81.5 && 2.81 & 640 & 347 & 90.2 && 3.23 & 539 & 353 & 87.3\\
hrnet32 & random  && 6.02 & 12547 & 2417 & 85.1 && 2.74 & 515 & 316 & 90.6 && 3.20 & 541 & 347 & 87.7\\
hrnet32 & worst  && 5.99 & 19419 & 2410 & 83.3 && 2.72 & 819 & 314 & 90.4 && 3.18 & 700 & 345 & 86.9\\
hrnet32 & max-distance  && 6.76 & 14252 & 2850 & 85.6 && 3.16 & 721 & 441 & 90.8 && 3.35 & 583 & 375 & 87.9\\
hrnet32 & round-robin  && 6.42 & 14279 & 2467 & 84.2 && 3.45 & 612 & 340 & 90.4 && 3.88 & 566 & 349 & 87.5\\
hrnet32 & worst with limit  && 6.000 & 12191 & 2414 & 84.7 && 2.72 & 496 & 311 & 90.5 && 3.18 & 528 & 347 & 87.6\\
\hdashline
hrnet32 & mean  && 6.22 & 14630 & 2511 & 84.1 && 2.93 & 634 & 345 & 90.5 && 3.34 & 576 & 353 & 87.5\\
hrnet32 & std  && 0.31 & 2596 & 170 & 1.5 && 0.31 & 123 & 49 & 0.20 && 0.27 & 64 & 11 & 0.35\\

\midrule
Swin-T & best  && 6.07 & 14853 & 2460 & 81.8 && 2.75 & 624 & 327 & 90.3 && 3.20 & 501 & 348 & 87.7\\
Swin-T & random  && 6.00 & 12710 & 2401 & 85.1 && 2.69 & 510 & 303 & 90.7 && 3.16 & 514 & 338 & 88.0\\
Swin-T & worst  && 5.94 & 19309 & 2369 & 83.4 && 2.68 & 798 & 300 & 90.5 && 3.16 & 704 & 341 & 87.1\\
Swin-T & max-distance  && 6.74 & 14277 & 2854 & 85.7 && 3.15 & 737 & 449 & 90.9 && 3.33 & 550 & 370 & 88.2\\
Swin-T & round-robin  && 6.37 & 14268 & 2438 & 84.3 && 3.40 & 595 & 325 & 90.6 && 3.84 & 534 & 339 & 87.9\\
Swin-T & worst with limit  && 5.9 & 12436 & 2390 & 84.8 && 2.68 & 492 & 302 & 90.7 && 3.16 & 503 & 340 & 88.0\\
\hdashline
Swin-T & mean  && 6.18 & 14642 & 2485 & 84.2 && 2.89 & 626 & 334 & 90.6 && 3.31 & 551 & 346 & 87.8\\
Swin-T & std  && 0.32 & 2478 & 184 & 1.4 && 0.31 & 122 & 57 & 0.20 && 0.27 & 77 & 12 & 0.39\\

\midrule
Swin-L & best  && 5.80 & 13876 & 2305 & 82.4 && 2.47 & 497 & 266 & 90.7 && 3.06 & 483 & 330 & 88.4\\
Swin-L & random  && 5.70 & 11958 & \textbf{2242} & 85.3 && 2.42 & 428 & \textbf{249} & 91.0 && 3.03 & 479 & 320 & 88.8\\
Swin-L & worst  && 5.66 & 18133 & 2242 & 83.7 && 2.41 & 671 & 251 & 90.8 && 2.99 & 620 & 314 & 88.1\\
Swin-L & max-distance  && 6.53 & 13107 & 2725 & \textbf{86.4} && 2.87 & 594 & 371 & \textbf{91.2} && 3.11 & 498 & 340 & \textbf{88.9} \\
Swin-L & round-robin  && 6.11 & 13639 & 2305 & 84.5 && 3.12 & 490 & 261 & 90.9 && 3.70 & 504 & 320 & 88.6\\
Swin-L & worst with limit  && \textbf{5.67} & \textbf{11565} & 2245 & 85.0 &&\textbf{2.41} & \textbf{422} & 250 & 90.9 && \textbf{2.99} & \textbf{461} & \textbf{315} & 88.7\\
\hdashline
Swin-L & mean  && 5.91 & 13713 & 2344 & 84.5 && 2.62 & 517 & 275 & 90.9 && 3.15 & 507 & 323 & 88.6\\
Swin-L & std  && 0.35 & 2351 & 189 & 1.4 && 0.30 & 98 & 48 & 0.17 && 0.27 & 57 & 10 & 0.29\\

\bottomrule
\end{tabularx}
\caption{Results on MIST using an IoU threshold of 85\%. NCI: normalised clicks per image, NFO: number of failed objects, NFI: number of failed images. All reported models are trained on COCO+LVIS.}
\label{tab:new_strat}
\end{table*}

\section{Extended Ablations}

Here, we extend the ablation experiments performed in \secAblations to additional datasets. Tab~\ref{tab:ablations_suppl} and Tab.~\ref{tab:single_inst_ablations} report the results of the ablation experiments on additional multi-instance and single-instance datasets respectively. As it can be seen from these experiments, our final model with spaio-temporal positional encoding consistently outperforms other variants, and is robust towards different task settings. Although, as stated in \secAblations, the impact of the spatial embedding seems to be less significant compared to the temporal counterpart in Tab~\ref{tab:ablations_suppl}, they are still important for reducing the overall number of clicks especially in the single-instance setting (ref Tab.~\ref{tab:single_inst_ablations}).

\section{Runtime and Memory Analysis}

\begin{figure*}
    \centering
    \begin{subfigure}[b]{0.48\textwidth}
        \includegraphics[width=\textwidth]{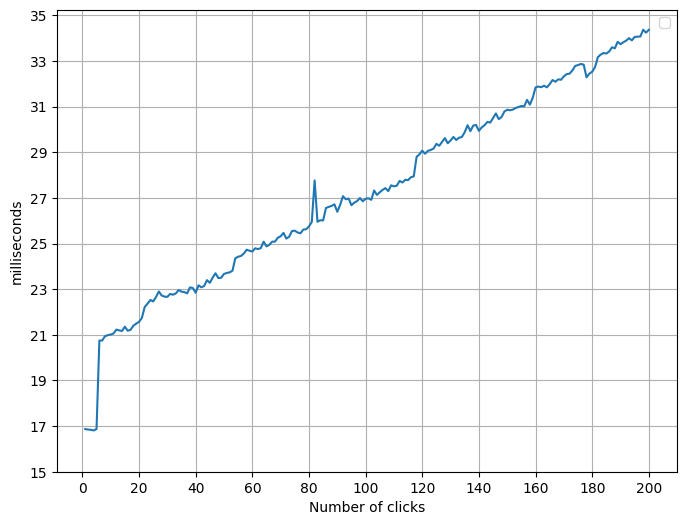}
        \caption{Runtime Analysis}
        \label{subfig:runtime}
    \end{subfigure}
    \begin{subfigure}[b]{0.48\textwidth}
        \includegraphics[width=\textwidth]{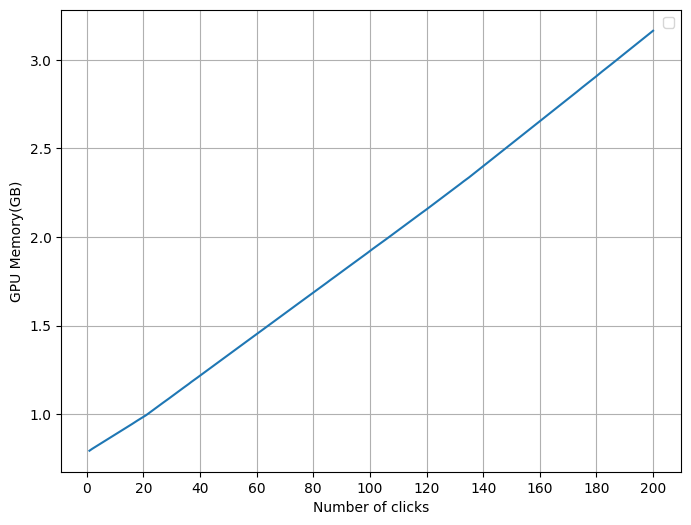}
        \caption{Memory Analysis}
        \label{subfig:mem}
    \end{subfigure}
    \caption{Runtime and memory scaling with respect to the number of clicks for the interactive transformer.}
    \label{fig:runtime_mem}
\end{figure*}

\begin{table*}[t!]
\setlength{\tabcolsep}{1pt} %
\centering
\small
\begin{tabularx}{\textwidth}{p{4.8cm}p{0.1cm}YYYp{0.1cm}YYYp{0.1cm}YYY}
\toprule%
&&\multicolumn{3}{c}{COCO}&&\multicolumn{3}{c}{SBD} && \multicolumn{3}{c}{DAVIS17}\\
\cmidrule{3-5}\cmidrule{7-9}\cmidrule{11-13}
  && NCI $\downarrow$ & NFO $\downarrow$& NFI $\downarrow$ && NCI $\downarrow$& NFO $\downarrow$& NFI $\downarrow$ && NCI $\downarrow$& NFO $\downarrow$& NFI $\downarrow$\\
\midrule
\abbrev{} (Swin-T) && \textbf{6.06} & \textbf{12997} & \textbf{2458} && \textbf{2.72} & \textbf{557} & \textbf{329} && \textbf{3.20} & \textbf{541} & \textbf{356} \\
- static background queries && 6.18 & 14436 & 2548 && 2.79 & 639 & 354 && 3.33 & 625 & 393 \\
- Transformer decoder && 6.34 & 14504 & 2652 && 2.90 & 657 & 384 && 3.24 & 582 & 371 \\
\midrule
- temporal positional encoding && 6.42 & 14729 & 2704 && 2.94 & 682 & 402 && 3.35 & 617 & 388 \\
- spatial positional encoding && 6.32 & 14506 & 2632 && 2.90 & 671 & 395 && 3.24 & 569 & 370 \\
- spatio-temporal positional encoding && 6.23 & 13552 & 2569 && 2.86 & 608 & 376 && 3.34 & 587 & 379 \\
\bottomrule
\end{tabularx}
\caption{Ablation on the network design choices, always relative to the top line. NCI: normalised clicks per image, NFO: number of failed objects, NFI: number of failed images. All reported models are trained on COCO+LVIS.}
\label{tab:ablations_suppl}
\end{table*}

\begin{table*}[t!]
\setlength{\tabcolsep}{1pt} %
\centering%
\footnotesize%
\begin{tabularx}{\textwidth}{p{4.2cm}p{0.1cm}YYp{0.1cm}YYp{0.1cm}YYp{0.1cm}YYp{0.1cm}YY}
\toprule%
&&\multicolumn{2}{c}{GrabCut~\cite{Rother04SIGGRAPH}}&&\multicolumn{2}{c}{Berkeley~\cite{mcguinness2010berkeley}} && \multicolumn{2}{c}{SBD~\cite{hariharan2011iccv}} && \multicolumn{2}{c}{COCO MVal}&& \multicolumn{2}{c}{DAVIS~\cite{Perazzi16CVPR}}\\
\cmidrule{3-4}\cmidrule{6-7}\cmidrule{9-10}\cmidrule{12-13}\cmidrule{15-16}
&& @85 $\downarrow$ & @90 $\downarrow$&& @85 $\downarrow$& @90 $\downarrow$&& @85 $\downarrow$& @90 $\downarrow$&& @85 $\downarrow$& @90  $\downarrow$&& @85 $\downarrow$& @90 $\downarrow$\\
\midrule
\abbrev{} (Swin-T) && 1.56 & 1.64 && 1.38 & 2.06 && \textbf{3.83} & \textbf{6.39} && \textbf{2.27} & 3.28 &&\textbf{ 3.75} & 5.19 \\
- static background queries && 1.64 & 1.68 && 1.35 & \textbf{1.87} && 3.92 & 6.51 && 2.31 & \textbf{3.21} && 3.84 & \textbf{5.15}\\
- Transformer decoder && 1.64 & 1.76 && \textbf{1.32} & 2.28 && 4.18 & 6.89 && 2.40 & 3.50 && 3.77 & 5.33 \\
\midrule
- temporal positional encoding && \textbf{1.52} & 1.64 && 1.51 & 2.27 && 4.17 & 6.89 && 2.42 & 3.48 && 4.04 & 5.43 \\
- spatial positional encoding && 1.76 & 1.86 && 1.36 & 2.41 && 4.19 & 6.89 && 2.44 & 3.45 && 3.84 & 5.28\\
- spatio-temporal positional encoding && 1.56 & \textbf{1.62} && 1.34 & 2.10 && 3.99 & 6.63 && 2.28 & 3.24 && 4.06 & 5.38 \\
\bottomrule
\end{tabularx}
\caption{Ablation on network design choice, on single-instance segmentation datasets, always relative to the top line.}
\label{tab:single_inst_ablations}
\end{table*}

As discussed in \secMethod, \abbrev{} translates each click into a query to our interactive transformer module. Hence, the number of queries processed by the transformer increases over time during the iterative refinement process. In Figure~\ref{fig:runtime_mem}, we analyze the impact of such a growing query pool in terms of runtime and GPU memory consumed during inference. Both the runtime and the memory increases as the transformer receives more queries, but the scale-up is quite slow and falls within a reasonable limit for practical usage. As shown in Fig.~\ref{subfig:runtime} and Fig.~\ref{subfig:mem}, the runtime increases from 17ms to 34ms as the number of clicks increases from 1 to 200, and the memory used increases from around 800MB to 3.2GB. For a large scale dataset like COCO with an average of 7.3 instances per image, \abbrev{} would need about 47 queries (since NCI is ~6.4) in the final refinement step and hence the average maximum runtime for a refinement step would be about 23.5ms. The values reported for both of these experiments in Fig.~\ref{fig:runtime_mem} are an average over the entire GrabCut dataset on an Nvidia 3090 GPU with 24GB of memory.

\section{Refinement Analysis}

\begin{figure*}
    \centering
    \begin{subfigure}[b]{0.48\textwidth}
        \includegraphics[width=\textwidth,trim=65 40 90 50, clip]{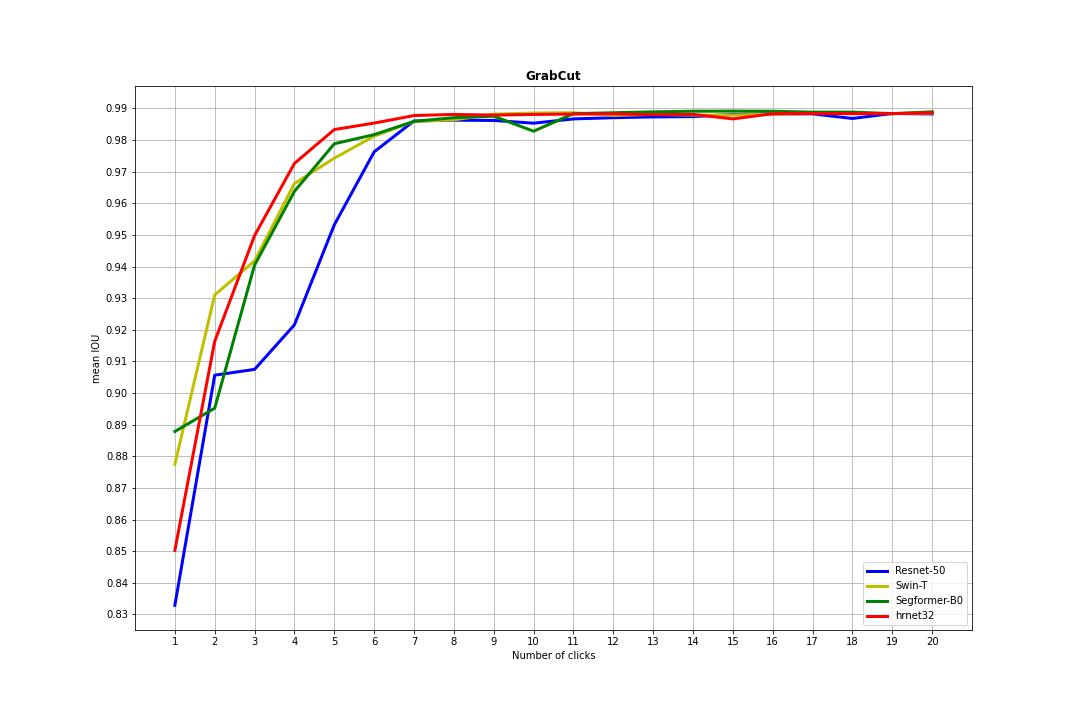}
    \end{subfigure}
    \begin{subfigure}[b]{0.48\textwidth}
        \includegraphics[width=\textwidth,trim=65 40 90 50, clip]{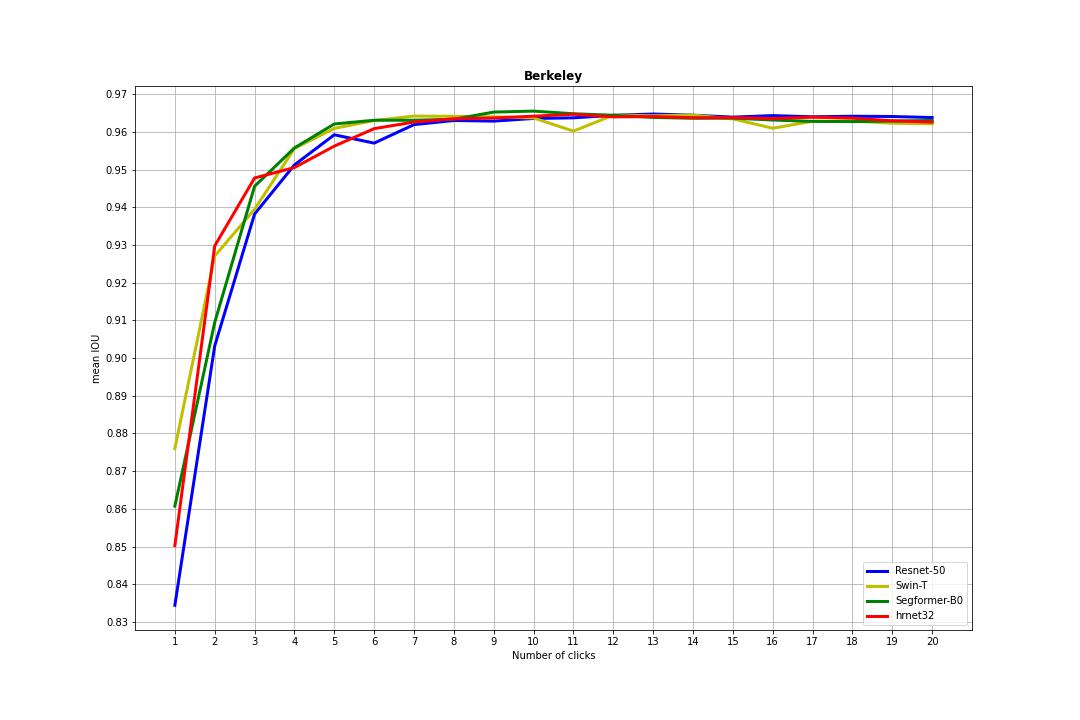}
    \end{subfigure}
    \begin{subfigure}[b]{0.48\textwidth}
        \includegraphics[width=\textwidth,trim=65 40 90 50, clip]{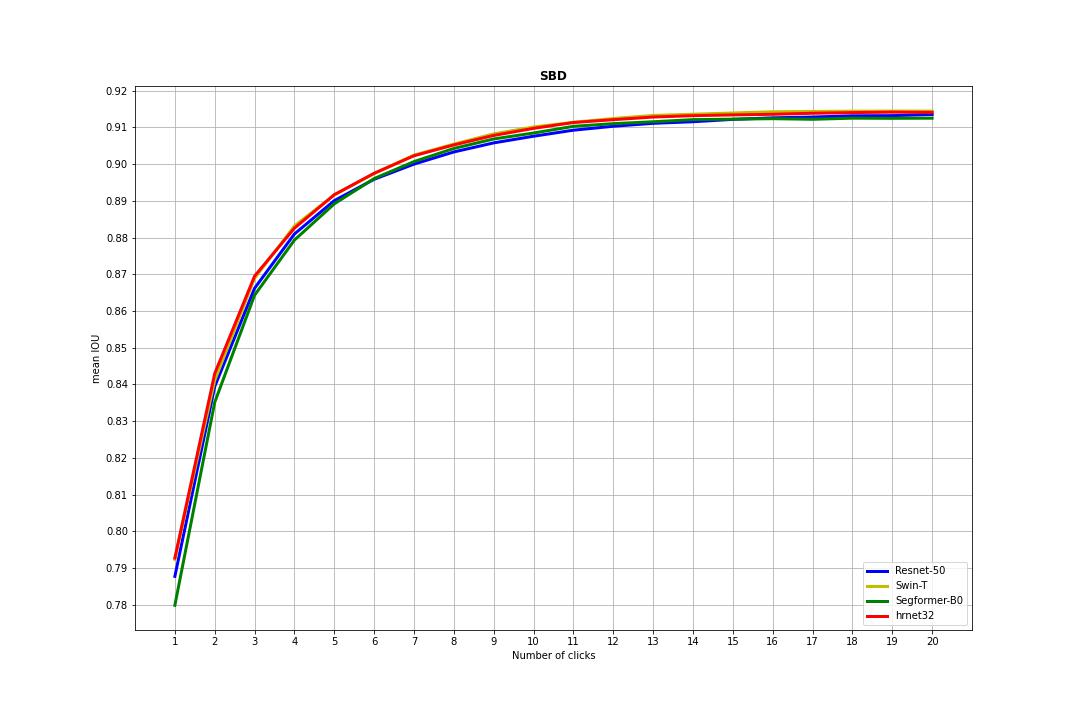}
    \end{subfigure}
    \begin{subfigure}[b]{0.48\textwidth}
        \includegraphics[width=\textwidth,trim=65 40 90 50, clip]{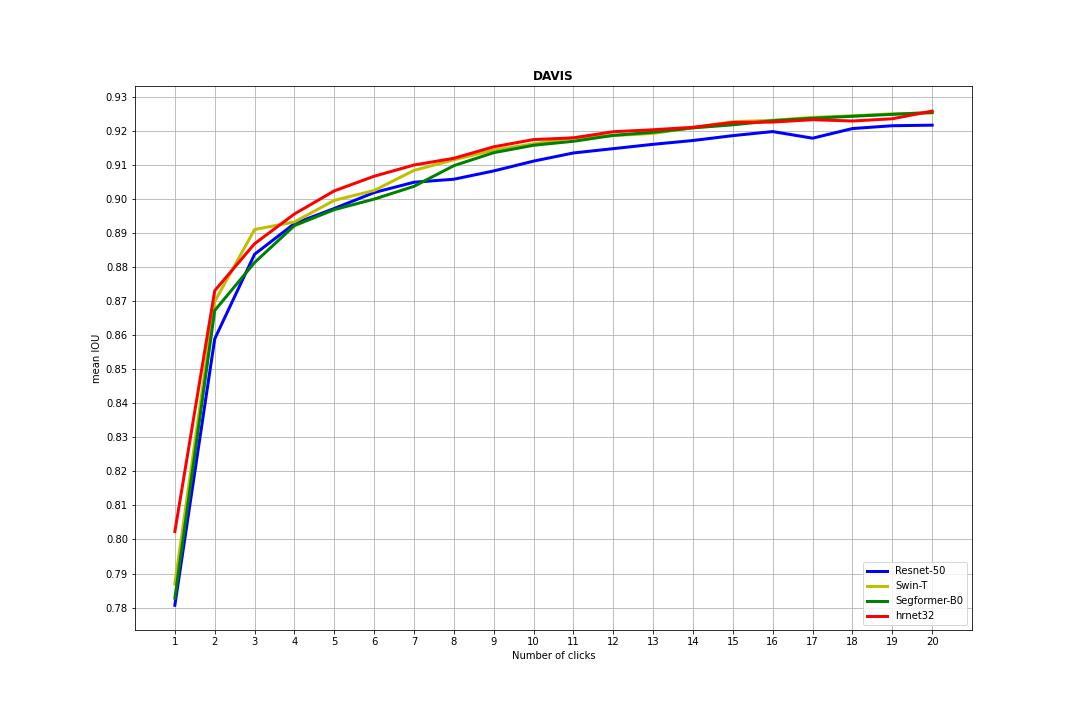}
    \end{subfigure}
    \caption{IoU vs. number of clicks for multiple single-instance datasets.}
    \label{fig:ref_quality}
\end{figure*}

In this section, we analyze the refinement quality of different variants of \abbrev{} for the single-instance setting.
Fig.~\ref{fig:ref_quality} plots change in instance segmentation quality after each refinement iteration on various single-instance datasets. \abbrev{} can achieve a high segmentation quality with very few clicks and can further refine the instances very well with additional clicks. Eg. for GrabCut, \abbrev{} achieves 84\% IoU on average with just one click, and then refines them to close to 100\% IoU. 

\section{Annotation Tool}

For using \abbrev{} in practice, we build a click based annotation tool that can perform multi-instance interactive segmentation. Our tool is built using the python based GUI toolkit \textit{Tkinter}, and is based on the RITM~\cite{SofiiukArxiv21} annotation tool. The \abbrev{} annotation tool supports addition and deletion of instances within an image, and also allows a user to switch back and forth between instances to perform mask refinement. To get a glimpse of our tool, please watch the video on the project page.

It should be noted that this tool is only a prototype and cannot be seen as a proper tool that was optimized for the best possible user experience. 
Many improvements could be thought of, \eg one could optimize the switching between objects by right-clicking on existing masks and keyboard shortcuts could be included for actions such as creating a new object.
We could also easily extend the tool with additional functionalities such as the removal of existing clicks, since this is supported out of the box by \abbrev{}.
A detailed exploration of this design space is outside of our expertise and the scope of this paper.

\section{Qualitative Results}
\newcommand{\im}[3]{\begin{tikzpicture}
        \node[inner sep=0pt] (picture) {\includegraphics[width=0.246\linewidth,trim={#2},clip]{#1}};
        \node[above right, fill=white] at (picture.south west) {\footnotesize #3};
    \end{tikzpicture}}

\newcommand{\imline}[8]{
        \includegraphics[width=0.246\linewidth,trim={#2},clip]{figures/suppl/qualitative/#1/iter_0.jpg}\hfill%
        \im{figures/suppl/qualitative/#1/#3}{#2}{#4}\hfill%
        \im{figures/suppl/qualitative/#1/#5}{#2}{#6}\hfill%
        \im{figures/suppl/qualitative/#1/#7}{#2}{#8}}
        
\begin{figure*}
    \centering
    \imline{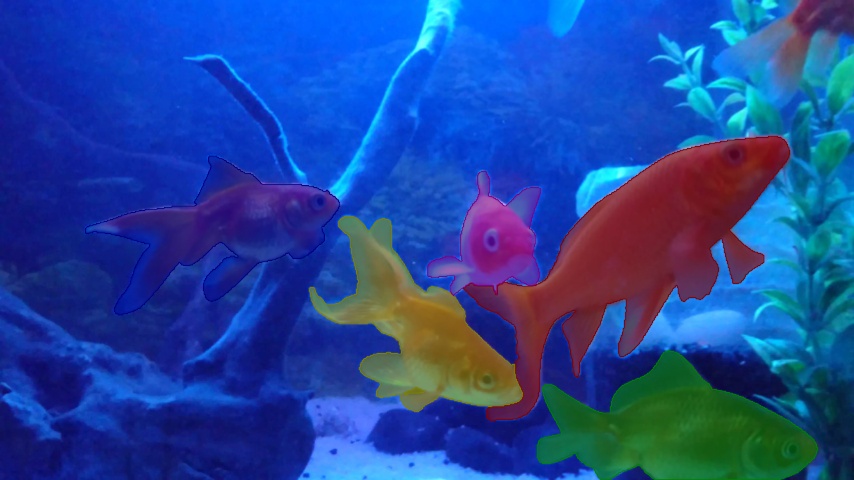}{0cm 0cm 0cm 1.7cm}{iter_5.jpg}{80.52}{iter_15.jpg}{90.18}{iter_25.jpg}{92.70}
    
    \imline{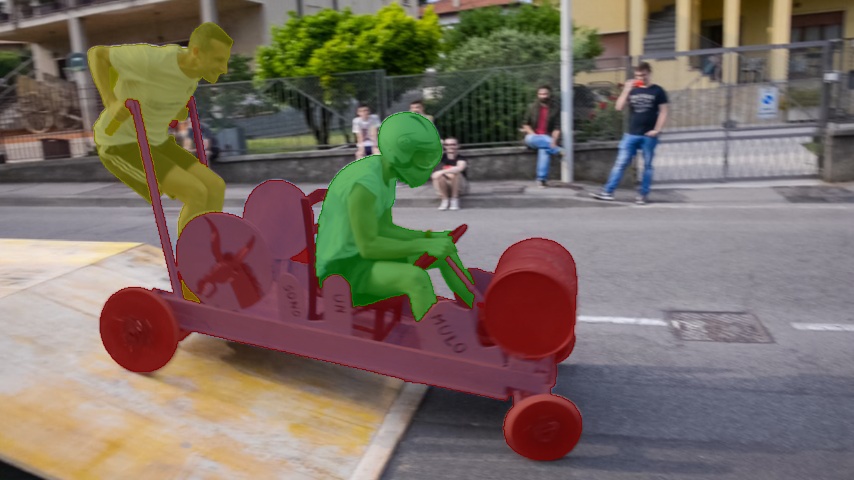}{0cm 0cm 0cm 0cm}{iter_3.jpg}{91.89}{iter_15.jpg}{92.96}{iter_25.jpg}{93.08}

    \imline{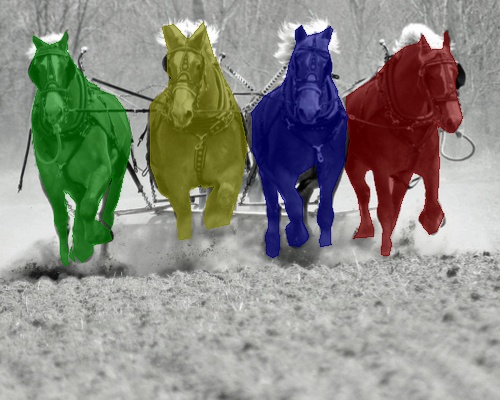}{0cm 2.7cm 0cm 0cm}{iter_4.jpg}{84.43}{iter_12.jpg}{88.42}{iter_20.jpg}{89.44}

    \imline{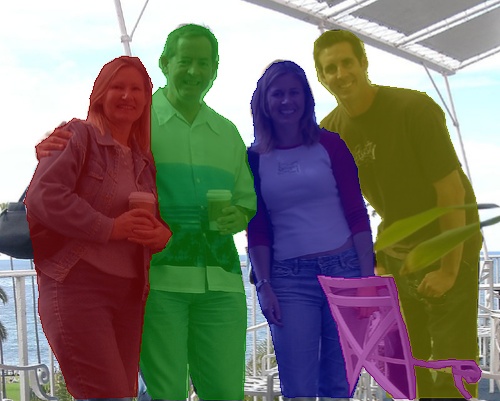}{0cm 0cm 0cm 0cm}{iter_5.jpg}{89.78}{iter_15.jpg}{93.74}{iter_25.jpg}{94.06}

    \imline{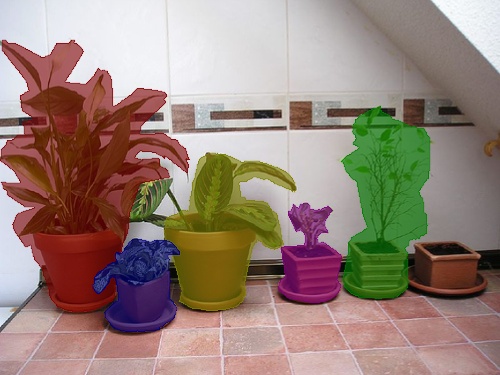}{0cm 0cm 0cm 1.5cm}{iter_5.jpg}{80.98}{iter_15.jpg}{89.63}{iter_25.jpg}{92.30}

    \imline{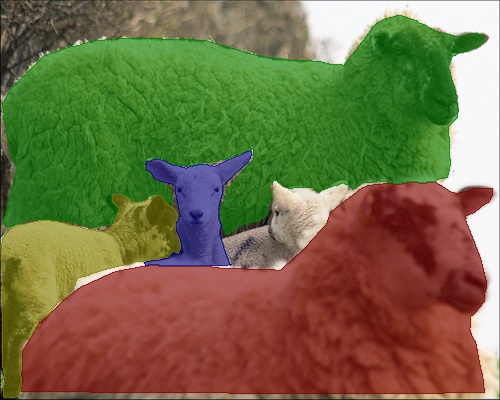}{0cm 0cm 0cm 0cm}{iter_4.jpg}{91.80}{iter_12.jpg}{92.65}{iter_20.jpg}{93.08}

    \imline{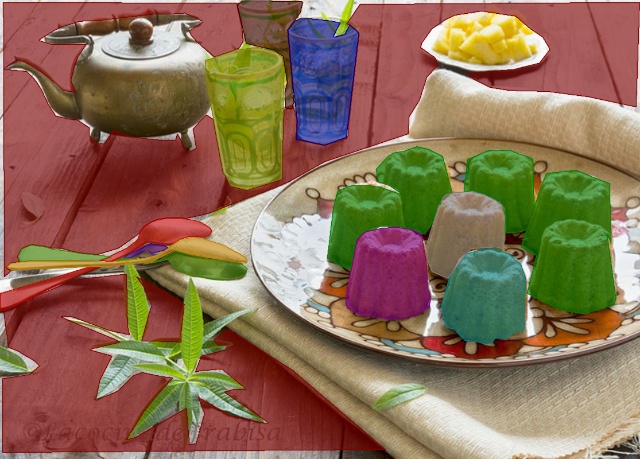}{0cm 0cm 0cm 0cm}{iter_12.jpg}{66.04}{iter_36.jpg}{77.42}{iter_60.jpg}{83.71}

    \begin{subfigure}[b]{0.246\linewidth}%
    \caption{Ground truth}%
    \end{subfigure}\hfill%
    \begin{subfigure}[b]{0.246\linewidth}%
    \caption{$\tau=1$}%
    \end{subfigure}\hfill%
    \begin{subfigure}[b]{0.246\linewidth}%
    \caption{$\tau=3$}%
    \end{subfigure}\hfill%
    \begin{subfigure}[b]{0.246\linewidth}%
    \caption{$\tau=5$}%
    \end{subfigure}%
    \caption{Qualitative examples based on our automatic random click sampling strategy. We show the ground truth and how the segmentation looks after a click budge of $\tau * |\mathcal{O}|$. For $\tau = 1$ we click on each object exactly once. The bottom left corner of each image shows the average IoU.}
    \label{fig:success}
\end{figure*}

\begin{figure*}
    \centering
    \imline{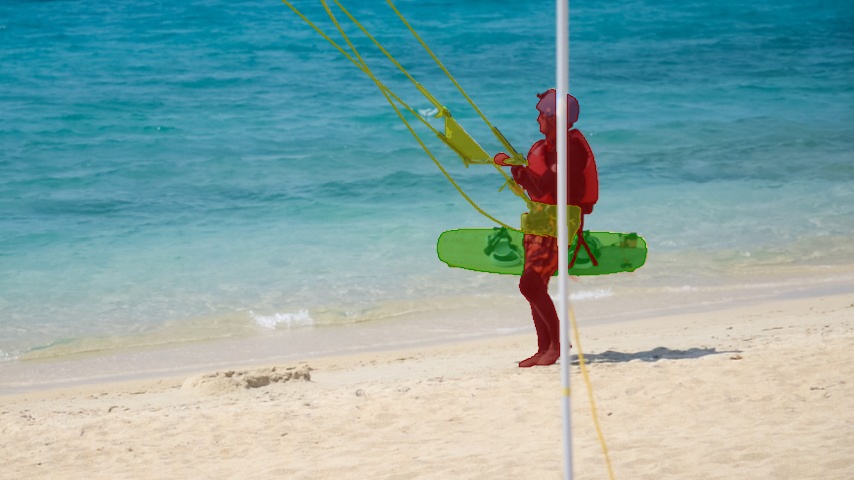}{5cm 0cm 1cm 1cm}{iter_3.jpg}{38.99}{iter_9.jpg}{66.47}{iter_15.jpg}{68.38}

    \begin{subfigure}[b]{0.246\linewidth}%
    \caption{Ground truth}%
    \end{subfigure}\hfill%
    \begin{subfigure}[b]{0.246\linewidth}%
    \caption{$\tau=1$}%
    \end{subfigure}\hfill%
    \begin{subfigure}[b]{0.246\linewidth}%
    \caption{$\tau=3$}%
    \end{subfigure}\hfill%
    \begin{subfigure}[b]{0.246\linewidth}%
    \caption{$\tau=5$}%
    \end{subfigure}%
    \caption{A qualitative example of a negative result.
    Even though both the board and the ropes of the kite are segmented badly, the board can be recovered with a few additional clicks.
    After a total of 15 clicks, the refinement is not able to segment the ropes though. Given that the refinement clicks are sampled based on a maximum distance transform, no clicks are sampled for the very thin structure, even though \abbrev{} might actually be able to segment such structures.}
    \label{fig:failure}
\end{figure*}
In Fig.~\ref{fig:success}, we show additional multi-instance segmentation results for sequential segmentation process using \abbrev{}.
Here we follow the \textit{random} strategy, where we first sample a single click per object, after which we iteratively select a random object to refine.
In most cases, \abbrev{} starts out with a high average IoU after a single click per object and the resulting masks are often arguably better than the corresponding ground truth segmentation, \eg row 3, 5, and 6.
Nevertheless, in most cases we can also adjust to arbitrary mistakes present in the ground truth annotations.
There are also some interesting failure cases such as the one shown in Fig.~\ref{fig:failure}, where \abbrev{} fails to capture the thin ropes of the kite.
Although \abbrev{} can segment fairly thin structures in practice, the automatic click sampling fails to sample the necessary additional clicks for \abbrev{} to segment the ropes in this particular case.
\end{document}